\documentclass{article}

\usepackage[final]{corl_2018}

\usepackage{csquotes}
\usepackage{graphicx}
\graphicspath{ {./images/} }
\usepackage{amsmath,amsfonts}
\usepackage[font=small,labelfont=bf]{caption}
\usepackage{multirow}
\DeclareMathOperator*{\argmax}{arg\,max}
\RequirePackage{algorithm}
\RequirePackage{algorithmic}

\usepackage{algorithm}
\usepackage{algorithmic}
\usepackage{wrapfig}
\usepackage{subcaption}
\usepackage{enumerate}
\usepackage{hyperref}

\newcommand{\ie}{i.e.\ }

\newcommand{\reals}{\mathbb{R}}

\newcommand{\bs}{\mathbf{s}}
\newcommand{\ba}{\mathbf{a}}

\newcommand{\states}{\mathcal{S}}
\newcommand{\actions}{\mathcal{A}}

\newcommand{\bellman}{\mathcal{E}}
\newcommand{\E}{\mathbb{E}}

\newcommand\img[1]%
  {\raisebox
    {\dimexpr-0.5\height+0.5ex}%
    {\includegraphics[width=0.95\textwidth]{#1}}%
  }
\newcounter{row}

\newenvironment{imgrows}[1][\textwidth]%
  {\begin{minipage}{#1}%
   \setcounter{row}{0}%
   \stepcounter{figure}%
  }%
  {\addtocounter{figure}{-1}%
   \end{minipage}%
  }
\newcommand\imgrow
  {\par\noindent
   \refstepcounter{row}%
   \makebox[1.5em][r]{(\alph{row})}\
  }

\newcommand\literallabel[2]{%
  \label{#1}
  \expandafter\gdef\csname literallabel_#1\endcsname{#2}
  #2
}
\newcommand\literalref[1]{$\csname literallabel_#1\endcsname$}

\newcommand{\numgrasps}{{580k}} %
\newcommand{\modelparams}{{1.2M}} %
\newcommand{\siteurl}{\url{https://goo.gl/ykQn6g}}

\title{QT-Opt: Scalable Deep Reinforcement Learning\\ for Vision-Based Robotic Manipulation} %

\author{
Dmitry Kalashnikov$^1$, Alex Irpan$^1$, Peter Pastor$^2$, Julian Ibarz$^1$,\\ \and
\textbf{Alexander Herzog$^2$, Eric Jang$^1$,
Deirdre Quillen$^3$, Ethan Holly$^1$,}\\ \and
\textbf{Mrinal Kalakrishnan$^2$, Vincent Vanhoucke$^1$, Sergey Levine$^{1,3}$} \\
\{dkalashnikov, alexirpan, julianibarz, ejang, eholly, vanhoucke, slevine\}@google.com,\\
\{peterpastor, alexherzog, kalakris\}@x.team, \{deirdrequillen\}@berkeley.edu
}
\begin{document}

\footnotetext[1]{Google Brain, United States}
\footnotetext[2]{X, Mountain View, California, United States}
\footnotetext[3]{University of California Berkeley, Berkeley, California, United States}
\setcounter{footnote}{3}

\maketitle

\begin{abstract}
In this paper, we study the problem of learning vision-based dynamic manipulation skills using a scalable reinforcement learning approach. We study this problem in the context of grasping, a longstanding challenge in robotic manipulation. In contrast to static learning behaviors that choose a grasp point and then execute the desired grasp, our method enables closed-loop vision-based control, whereby the robot continuously updates its grasp strategy based on the most recent observations to optimize long-horizon grasp success. To that end, we introduce QT-Opt, a scalable self-supervised vision-based reinforcement learning framework that can leverage over \numgrasps{} real-world grasp attempts to train a deep neural network Q-function with over \modelparams{} parameters to perform closed-loop, real-world grasping that generalizes to 96\% grasp success on unseen objects. Aside from attaining a very high success rate, our method exhibits behaviors that are quite distinct from more standard grasping systems: using only RGB vision-based perception from an over-the-shoulder camera, our method automatically learns regrasping strategies, probes objects to find the most effective grasps, learns to reposition objects and perform other non-prehensile pre-grasp manipulations, and responds dynamically to disturbances and perturbations.\footnote{Supplementary experiment videos can be found at \siteurl.}

\end{abstract}


\section{Introduction}
Manipulation with object interaction represents one of the largest open problems in robotics: intelligently interacting with previously unseen objects in open-world environments requires generalizable perception, closed-loop vision-based control, and dexterous manipulation. Reinforcement learning offers a promising avenue for tackling this problem, but current work on reinforcement learning tackles the problem of mastering individual skills, such as hitting a ball~\citep{peters_2008}, opening a door~\citep{kalakris11,yahya17}, or throwing~\citep{kth17}.
To meet the generalization demands of real-world manipulation, we focus specifically on scalable learning with off-policy algorithms, and study this question in the context of the specific problem of grasping.
While grasping restricts the manipulation problem, it still retains many of its largest challenges: a grasping system should be able to pick up previously unseen objects with reliable and effective grasps, while using realistic sensing and actuation. It thus serves as a microcosm of the larger robotic manipulation problem, providing a challenging and practically applicable model problem for experimenting with generalization and diverse object interaction.
Much of the existing work on robotic grasping decomposes the task into a sensing, planning, and acting stage: the robot first perceives the scene and identifies suitable grasp locations, then plans a path to those locations~\citep{zeng2018, juxi18, dexnet30_2017, platt_gpd_17}. This stands in contrast to the kinds of grasping behaviors observed in humans and animals, where the grasp is a dynamical process that tightly interleaves sensing and control at every stage~\citep{rodriguez2018icra, bohg2014}.
This kind of dynamic closed-loop grasping is likely to be much more robust to unpredictable object physics, limited sensory information (e.g., monocular camera inputs instead of depth), and imprecise actuation. A closed-loop grasping system trained for long-horizon success can also perform intelligent pre-grasping manipulations, such as pushing or repositioning objects for an easier grasp. However, a major challenge with closed-loop grasp control is that the sensorimotor loop must be closed on the visual modality, which is very difficult to utilize effectively with standard optimal control methods in novel settings.
\begin{wrapfigure}{r}{0.5\textwidth}
\vspace{-0.3cm}
 \includegraphics[width=0.5\textwidth]{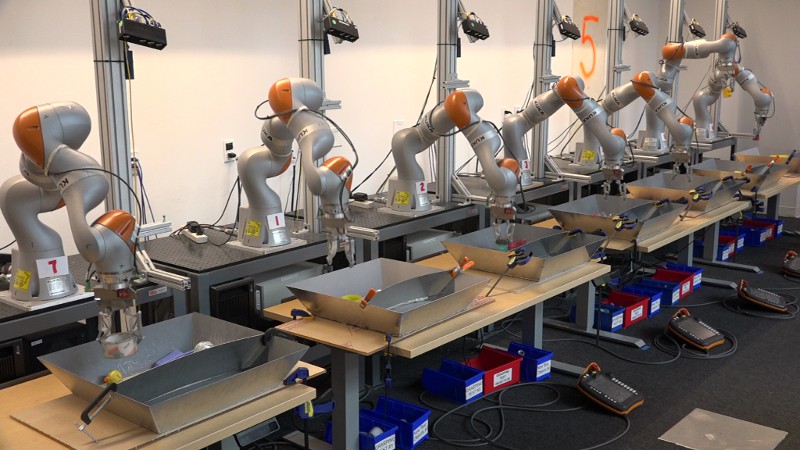}
\vspace{-0.4cm}
   \caption{Seven robots are set up to collect grasping episodes with autonomous self-supervision.}
\label{fig:teaser}
\vspace{-0.3cm}
\end{wrapfigure}
We study how off-policy deep reinforcement learning can acquire closed-loop dynamic visual grasping strategies, using entirely self-supervised data collection, so as to generalize to previously unseen objects at test time. The value of low-level end-effector movements is predicted directly from raw camera observations, and the entire system is trained using grasp attempts in the real world. While the principles of deep reinforcement learning have been known for decades~\citep{sutton98,tesauro94}, operationalizing them in a practical robotic learning algorithm that can generalize to new objects requires a stable and scalable algorithm and large datasets, as well as careful system design.

\begin{wrapfigure}{l}{0.5\textwidth}
\vspace{-0.3cm}
 \includegraphics[width=0.5\textwidth]{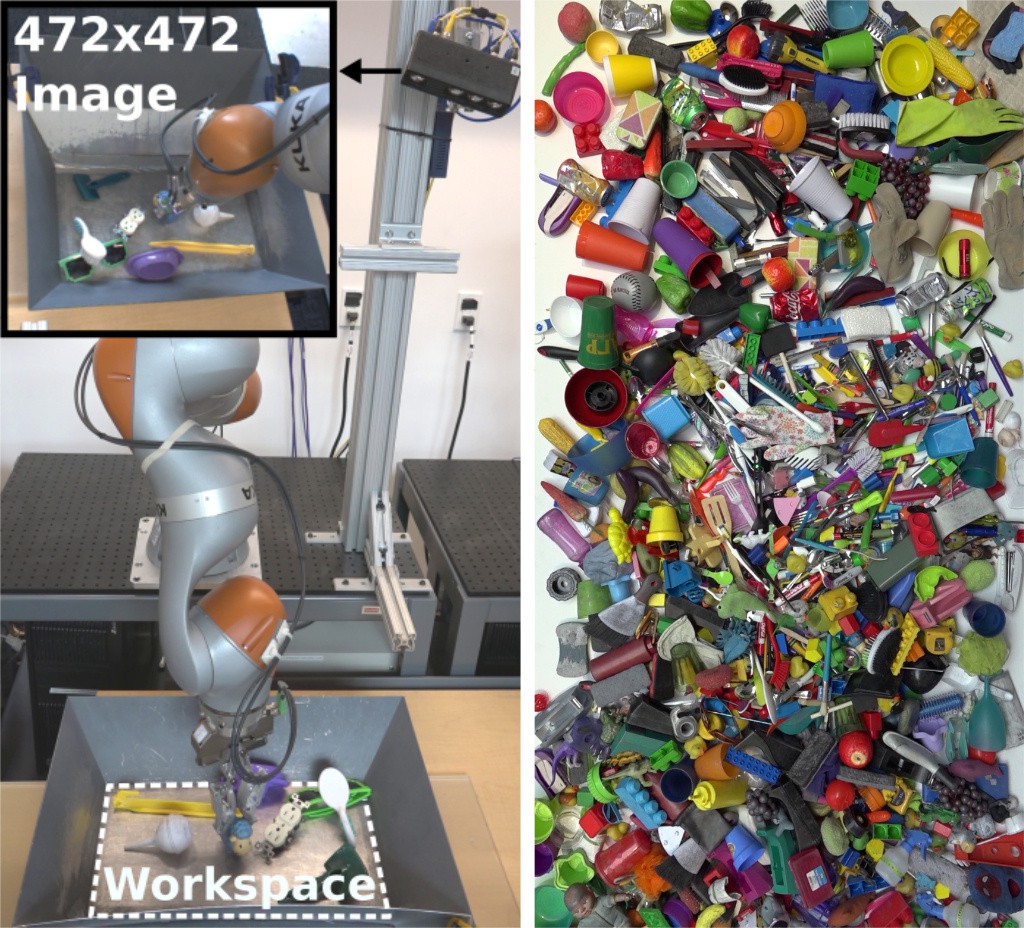}
\vspace{-0.4cm}
   \caption{Close-up of a robot cell in our setup (left) and about 1000 visually and physically diverse training objects (right). Each cell (left) consists of a KUKA LBR IIWA arm with a two-finger gripper and an over-the-shoulder RGB camera.}
\label{fig:robot_setup_and_objects}
\vspace{-0.3cm}
\end{wrapfigure}

The implementation in our experiments makes very simple assumptions: observations come from a monocular RGB camera located over the shoulder (see Fig.~\ref{fig:robot_setup_and_objects}), and actions consist of end-effector Cartesian motion and gripper opening and closing commands. The reinforcement learning algorithm receives a binary reward for lifting an object successfully, and no other reward shaping. This general set of assumptions makes the method feasible to deploy at large scale, allowing us to collect \numgrasps{} grasp attempts on 7 real robotic systems. Unlike most reinforcement learning tasks in the literature~\citep{machado17arcade,gym16}, the primary challenge in this task is not just to maximize reward, but to generalize effectively to previously unseen objects. This requires a very diverse set of objects during training. To make maximal use of this diverse dataset, we propose an off-policy training method based on a continuous-action generalization of Q-learning, which we call QT-Opt. Unlike other continuous action Q-learning methods~\citep{hafner11, lillicrap15}, which are often unstable due to actor-critic instability~\citep{duan16,deeprlthatmatters17}, QT-Opt dispenses with the need to train an explicit actor, instead using stochastic optimization over the critic to select actions and target values~\citep{gcg,quillen}. We show that even fully off-policy training can outperform strong baselines based on prior work, while a moderate amount of on-policy joint finetuning with offline data can improve performance to a success rate of 96\% on challenging, previously unseen objects.

Our experimental evaluation demonstrates the effectiveness of this approach both quantitatively and qualitatively. We show that our method attains a high success rate across a range of objects not seen during training, and our qualitative experiments show that this high success rate is due to the system adopting a variety of strategies that would be infeasible without closed-loop vision-based control: the learned policies exhibit corrective behaviors, regrasping, probing motions to ascertain the best grasp, non-prehensile repositioning of objects, and other features that are feasible only when grasping is formulated as a dynamic, closed-loop process.

\section{Related Work}

Reinforcement learning has been applied in the context of robotic control using both low-dimensional~\citep{peters_2008, kalakris11} and high-dimensional~\citep{hafner11, lillicrap15} function approximators, including with visual inputs~\citep{levine2015,yahya17}. However, all of these methods focus on learning narrow, individual tasks, and do not evaluate on broad generalization to large numbers of novel test objects. Real-world robotic manipulation requires broad generalization, and indeed much of the research on robotic grasping has sought to achieve such generalization, either through the use of grasp metrics based on first principles~\citep{weisz2012} or learning~\citep{lenz2015,bohg2014}, with the latter class of methods achieving some of the best results in recent years~\citep{platt_gpd_17,dexnet30_2017}. However, current grasping systems typically approach the grasping task as the problem of predicting a \emph{grasp pose}, where the system looks at the scene (typically using a depth camera), chooses the best location at which to grasp, and then executes an open-loop planner to reach that location~\citep{zeng2018, juxi18, dexnet30_2017, platt_gpd_17}. In contrast, our approach uses reinforcement learning with deep neural networks, which enables dynamic closed-loop control. This allows our policies to perform pre-grasp manipulation and respond to dynamic disturbances and, crucially, allows us to learn grasping in a generic framework that makes minimal assumptions about the task.

While most prior grasping methods operate in open-loop, a number of works have studied closed-loop grasping~\citep{yu2018icra,platt17,hausman17,levine16}. In contrast to these methods, which frame closed-loop grasping as a servoing problem, our method uses a general-purpose reinforcement learning algorithm to solve the grasping task, which enables long-horizon reasoning. In practice, this enables our method to autonomously acquire complex grasping strategies, some of which we illustrate in Section~\ref{sec:experiments}. Our method is also entirely self-supervised, using only grasp outcome labels that are obtained automatically by the robot. Several works have proposed self-supervised grasping systems~\citep{pinto16,levine16}, but to our knowledge, ours is the first to incorporate long-horizon reasoning via reinforcement learning into a generalizable vision-based system trained on self-supervised real-world data. Related to our work, \citet{zeng2018} recently proposed a Q-learning framework for combining grasping and pushing. Our method utilizes a much more generic action space, directly commanding gripper motion in 3D, and exhibits substantially better performance and generalization in our experiments. Finally, in contrast to many current grasping systems that utilize depth sensing~\cite{dexnet30_2017, morrison18} or wrist-mounted cameras~\cite{platt17, morrison18}, our method operates on raw monocular RGB observations from an over-the-shoulder camera, and the performance of our method indicates that effective learning can achieve excellent grasp success rates even with very rudimentary sensing.

\section{Overview}
\begin{wrapfigure}{rb}{0.5\textwidth}
\vspace{-0.5cm}
\begin{center}
 \includegraphics[width=0.5\textwidth]{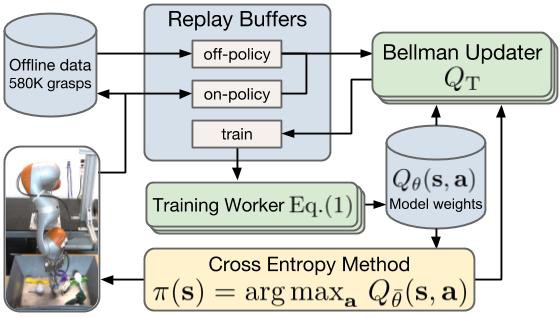}
\end{center}
   \vspace{-0.35cm}
   \caption{Our distributed RL infrastructure for QT-Opt (see Sec.~\ref{sec:qtopt-details}). State-action-reward tuples are loaded from an offline data stored and pushed from online real robot collection (see Sec.~\ref{sec:grasping_system}). Bellman update jobs sample transitions and generate training examples, while training workers update the Q-function parameters.}
\label{fig:distributed_infra_sketch}
\vspace{-0.2cm}
\end{wrapfigure}
Our closed-loop vision-based control framework is based on a general formulation of robotic manipulation as a Markov Decision Process (MDP)\footnote{While a partially observed (POMDP) formulation would be most general, we assume that the current observation provides all necessary information. In practice, the resulting policy still exhibits moderate robustness to occlusions, and a more general extension to recurrent policies and Q-functions would be straightforward.}. At each time step, the policy observes the image from the robot's camera (see Fig.~\ref{fig:robot_setup_and_objects}) and chooses a gripper command, as discussed in Section~\ref{sec:grasping_system}.
This task formulation is general and could in principle be applied to a wide range of robotic manipulation tasks. The grasping task is defined simply by providing a reward to the learner during data collection: a successful grasp results in a reward of $1$, and a failed grasp a reward of $0$. A grasp is considered successful if the robot holds an object above a certain height at the end of the episode.

The framework of MDPs provides a general and powerful formalism for such decision-making problems, but
learning in this framework can be challenging.
Generalization requires diverse data, but recollecting experience on a wide range of objects after every policy update is impractical, ruling out on-policy algorithms. Instead, we devise a scalable off-policy reinforcement learning framework based around a continuous generalization of Q-learning. While actor-critic algorithms are a popular approach in the continuous action setting, we found that a more stable and scalable alternative is to train only a Q-function, and induce a policy implicitly by maximizing this Q-function using stochastic optimization. We describe the resulting algorithm, which we call QT-Opt, in Section~\ref{sec:qtopt}, and describe its instantiation for robotic grasping in Section~\ref{sec:grasping_system}.
To handle the large datasets and networks in our approach, we devise a distributed collection and training system that asynchronously updates target values,
collects on-policy data, reloads off-policy data from past experiences, and trains the network on both data streams within a distributed optimization framework (see Fig.~\ref{fig:distributed_infra_sketch}).

\section{Scalable Reinforcement Learning with QT-Opt}
\label{sec:qtopt}

In this section, we describe the reinforcement learning algorithm that we use for our closed-loop vision-based grasping method. The algorithm is a continuous action version of Q-learning adapted for scalable learning and optimized for stability,
to make it feasible to handle large amounts of off-policy image data for complex tasks like grasping.

\subsection{Reinforcement Learning and Q-Learning}
\label{sec:rlql}

We first review the fundamentals of reinforcement learning and Q-learning, which we build on to derive our algorithm. We will use $\bs \in \states$ to denote the state, which in our case will include image observations (see Appendix~\ref{sec:appendix_state_action_reward} for details). $\ba \in \actions$ denotes the action, which will correspond to robot arm motion and gripper command. At each time step $t$, the algorithm chooses an action, transitions to a new state, and receives a reward $r(\bs_t,\ba_t)$. The goal in RL is to recover a policy that selects actions to maximize the total expected reward. One way to acquire such an optimal policy is to first solve for the optimal Q-function, which is sometimes referred to as the state-action value function. The Q-function specifies the expected reward that will be received after taking some action $\ba$ in some state $\bs$, and the optimal Q-function specifies this value for the optimal policy. In practice, we aim to learn parameterized Q-functions $Q_\theta(\bs,\ba)$, where $\theta$ might denote the weights in a neural network. We can learn the optimal Q-function by minimizing the Bellman error, given by
\begin{equation}
\literallabel{eq:bellman}{
\bellman(\theta) = \E_{(\bs,\ba,\bs') \sim p(\bs,\ba,\bs')} \left[ D \left(
Q_\theta(\bs,\ba) ,Q_T(\bs,\ba,\bs') \right)
\right],}
\end{equation}
where $Q_T(\bs,\ba,\bs') = r(\bs,\ba) + \gamma V(\bs')$ is a \emph{target value}, and $D$ is some divergence metric. We use the cross-entropy function for $D$, since total returns are bounded in $[0,1]$, which we found to be more stable than the standard squared difference (see Appendix~\ref{sec:appendix_ablations_sim}). The expectation is taken under the distribution over all previously observed transitions, and $V(\bs')$ is a target value. In our implementation, we use two target networks~\citep{hafner11,mnih2015,gu16} to improve stability, by maintaining two lagged versions of the parameter vector $\theta$, $\bar{\theta}_1$ and $\bar{\theta}_2$, where $\bar{\theta}_1$ is the exponential moving averaged version of $\theta$ with an averaging constant of 0.9999, and $\bar{\theta}_2$ is a lagged version of $\bar{\theta}_1$, which is lagged by about 6000 gradient steps. We then compute the target value according to \literallabel{eq:v}{$V(\bs') = \min_{i=1,2} Q_{\bar{\theta}_i}(\bs',\arg\max_{\ba'} Q_{\bar{\theta}_1}(\bs',\ba'))$}. This corresponds to a combination of Polyak averaging~\citep{polyak1992acceleration,lhph-ccdrl-16} and clipped double Q-learning~\citep{hasselt10,hasselt16,td3}, and we discuss this design decision further in Appendix~\ref{sec:appendix_ablations_sim}. Once the Q-function is learned, the policy can be recovered according to $\pi(\bs) = \arg\max_{\ba} Q_{\bar{\theta}_1} (\bs,\ba)$. Practical implementations of this method collect samples from environment interaction and then perform off-policy training on all samples collected so far~\citep{hafner11,mnih2015,gu16}. For large-scale learning problems of the sort tackled in this work, a parallel asynchronous version of this procedure substantially improves our ability to scale up this process, as discussed in Section~\ref{sec:rl_system}.

\subsection{QT-Opt for Stable Continuous-Action Q-Learning}
\label{sec:qtopt-details}

Q-learning with deep neural network function approximators provides a simple and practical scheme for RL with image observations, and is amenable to straightforward parallelization. However, incorporating continuous actions, such as continuous gripper motion in our grasping application, poses a challenge for this approach. Prior work has sought to address this by using a second network that amortizes the maximization~\citep{hafner11,lillicrap15}, or constraining the Q-function to be convex in $\ba$, making it easy to maximize analytically~\citep{gu16,amos2017icnn}. Unfortunately, the former class of methods are notoriously unstable~\citep{deeprlthatmatters17}, which makes it problematic for large-scale RL tasks where running hyperparameter sweeps is prohibitively expensive. Action-convex value functions are a poor fit for complex manipulation tasks such as grasping, where the Q-function is far from convex in the input. For example, the Q-value may be high for actions that reach toward objects, but low for the gaps between objects.

We therefore propose a simple and practical alternative that maintains the generality of non-convex Q-functions while avoiding the need for a second maximizer network. The image $\bs$ and action $\ba$ are inputs into our network, and the $\argmax$ in Equation~(\ref{eq:bellman}) is evaluated with a stochastic optimization algorithm that can handle non-convex and multimodal optimization landscapes, similarly to \citep{gcg} and \citep{quillen}. Let $\pi_{\bar{\theta}_1}(\bs)$ be the policy implicitly induced by the Q-function $Q_{\bar{\theta}_1}(\bs,\ba)$. We can recover Equation~(\ref{eq:bellman}) by substituting the optimal policy $\pi_{\bar{\theta}_1}(\bs) = \arg\max_{\ba} Q_{\bar{\theta}_1}(\bs,\ba)$ in place of the $\arg\max$ argument to the target Q-function. In our algorithm, which we call QT-Opt, $\pi_{\bar{\theta}_1}(\bs)$ is instead evaluated by running a stochastic optimization over $\ba$, using $Q_{\bar{\theta}_1}(\bs,\ba)$ as the objective value. We use the cross-entropy method (CEM) to perform this optimization, which is easy to parallelize and moderately robust to local optima for low-dimensional problems~\citep{rk-cem-04}. CEM is a simple derivative-free optimization algorithm that samples a batch of $N$ values at each iteration, fits a Gaussian distribution to the best \(M < N\) of these samples, and then samples the next batch of \(N\) from that Gaussian. In our implementation, we use \(N=64\) and \(M=6\), and perform two iterations of CEM. This is used both to compute targets at training time, and to choose actions in the real world.

\subsection{Distributed Asynchronous QT-Opt}
\label{sec:rl_system}

Learning vision based policies with reinforcement learning that generalizes over new scenes and objects requires large amounts of diverse data, in the same way that learning to generalize on complex vision tasks with supervised learning requires large datasets. For the grasping task in our experiments, we collected over \numgrasps{} grasps over the course of several weeks across 7 robots. To effectively train on such large and diverse RL dataset, we develop a distributed, asynchronous implementation of QT-Opt. Fig.~\ref{fig:distributed_infra_sketch} summarizes the system.
Transitions are stored in a distributed replay buffer database, which both loads historical data from disk and can accept online data from live ongoing experiments across multiple robots. The data in this buffer is continually labeled with target Q-values by using a set of 1000 ``Bellman updater'' jobs, which carry out the CEM optimization procedure using the current target network, and then store the labeled samples in a second training buffer, which operates as a ring buffer. One consequence of this asynchronous procedure is that some samples in the training buffer are labeled with lagged versions of the Q-network. This is discussed in more detail in the supplement, in Appendix~\ref{sec:appendix_bellman_update}. Training workers pull labeled transitions from the training buffer randomly and use them to update the Q-function. We use 10 training workers, each of which compute gradients which are sent asynchronously to parameter servers. We found empirically that a large number of gradient steps (up to 15M) were needed to train an effective Q-function due to the complexity of the task and large size of the dataset and model. Full details of the system design are provided in Appendix~\ref{sec:appendix_distributed_rl_infra}.

\section{Dynamic Vision-Based Grasping}
\label{sec:grasping_system}

In this section, we discuss how QT-Opt can be applied to enable dynamic vision-based grasping. An illustration of our grasping setup is shown in Fig.~\ref{fig:teaser}. The task requires a policy that can locate an object, position it for grasping (potentially by performing pre-grasp manipulations), pick up the object, potentially regrasping as needed, raise the object, and then signal that the grasp is complete to terminate the episode. To
enable self-supervised grasp labeling
in the real world, the reward only indicates whether or not an object was successfully picked up. This represents a fully end-to-end approach to grasping: no prior knowledge about objects, physics, or motion planning is provided to the model aside from the knowledge that it can extract autonomously from the data.

\vspace{-0.1in}
\paragraph{MDP for grasping.} The state observation $\bs \in \states$ includes the robot's current camera observation, an RGB image with a resolution of 472x472, recorded from an over-the-shoulder monocular camera (see Fig.~\ref{fig:teaser}).
We also found it beneficial to include the current status of the gripper in the state, which is a binary indicator of whether the gripper is open or closed, as well as the vertical position of the gripper relative to the floor (see comparisons in Appendix~\ref{sec:appendix_ablations_sim}). The action $\ba \in \actions$ consists of a vector in Cartesian space $\mathbf{t} \in \reals^3$ indicating the desired change in the gripper position, a change in azimuthal angle encoded via a sine-cosine encoding $\mathbf{r} \in \reals^2$, binary gripper open and close commands $g_\text{open}$ and $g_\text{close}$, and a termination command $e$ that ends the episode, such that ${\ba = (\mathbf{t}, \mathbf{r}, g_\text{open}, g_\text{close}, e)}$. Full details of the grasping MDP formulation are provided in Appendix~\ref{sec:appendix_state_action_reward}.

\vspace{-0.1in}
\paragraph{Reward function.}
The reward is 1 at the end of the episode if the gripper contains an object and is above a certain height, and 0 otherwise. Success is determined by using a background subtraction test after dropping the picked object, as discussed in Appendix~\ref{sec:grasp_success}. Note that this type of delayed and sparse reward function is generally quite challenging for reinforcement learning systems,
but it is also the most practical reward function for automated self-supervision. To encourage the robot to grasp more quickly, we also provide a small penalty $r(\bs_t,\ba_t) = -0.05$ for all time steps prior to termination, when the model either emits the termination action or exceeds the maximum number of time steps (20). This penalty may in principle result in target values outside of $[0,1]$, though we found empirically that this does not happen.

\vspace{-0.1in}
\paragraph{Q-Function representation.} The Q-function $Q_{\bar{\theta}_1}(\bs,\ba)$ is represented in our system by a large convolutional neural network with \modelparams{} parameters, where the image is provided as an input into the bottom of the convolutional stack, and the action, gripper status, and distance to floor are fed into the middle of the stack. The full neural network architecture is discussed in Appendix~\ref{sec:appendix_arch}.

\vspace{-0.1in}
\paragraph{Data collection.}
In order to enable our model to learn generalizable strategies that can pick up new objects, perform pre-grasp manipulation, and handle dynamic disturbances with vision-based feedback, we must train it on a sufficiently large and diverse set of objects. Collecting such data in a single on-policy training run would be impractical. Our off-policy QT-Opt algorithm makes it possible to pool experience from multiple robots and multiple experiments. The full dataset used to train our final model was collected over the course of four months, with a total of about 800 robot hours. This data was collected during multiple separate experiments, and each experiment reused the data from the previous one. This reduces our ability to provide rigidly controlled experimental results in the real-world system, but we provide more rigidly controlled results in simulation in the supplement, in Appendix~\ref{sec:appendix_ablations_sim}.
Since a completely random initial policy would produce a very low success with such an unconstrained action space, we use a weak scripted exploration policy to bootstrap data collection. This policy is randomized, but biased toward reasonable grasps, and achieves a success rate around 15-30\%. We switched to using the learned QT-Opt policy once it reached a success rate of 50\%. The scripted policy is described in the supplementary material, in Appendix~\ref{sec:appendix_exploration}. 
Data was collected with 7 LBR IIWA robots, with 4-10 training objects per robot. The objects were replaced every 4 hours during business hours, and left unattended at night and on weekends. The objects used during testing were distinct from those in the training data.

\section{Experimental Results}
\label{sec:experiments}

Our experiments evaluate our learned closed-loop vision-based grasping system to answer the following research questions: (1) How does our method perform, quantitatively, on new objects that were never seen during training? (2) How does its performance compare to a previously proposed self-supervised grasping system that does not explicitly optimize for long-horizon grasp success? (3) What types of manipulation strategies does our method adopt, and does it carry out meaningful, goal-directed pre-grasp manipulations? 
(4) How do the various design choices in our method affect its performance? The first two questions are addressed through a set of rigorous real-world quantitative experiments, which we discuss in Section~\ref{sec:quantitative}, question (3) is addressed through qualitative experiments, which are discussed in Section~\ref{sec:qualitative} and shown in the supplementary video and online,
and the last question is addressed through a detailed set of ablation studies in both simulation and the real world, which are discussed in Appendix~\ref{sec:appendix_ablations_sim} and \ref{sec:appendix_ablations_real}. The experiments in the appendices also study the impact of dataset size and off-policy training on final performance.

\subsection{Quantitative Performance Evaluation}
\label{sec:quantitative}

In this section, we present a quantitative evaluation of our grasping system. The physical setup for each robot is shown in Fig.~\ref{fig:teaser} (left): the robots are tasked with grasping objects in a bin, using an over-the-shoulder RGB camera and no other sensing.\footnote{Though some of the figures show a wrist-mounted camera, this camera is not used in the experiments.} We use two separate evaluation protocols, which use challenging objects that were not seen at training time. In the first protocol, each of the 7 robots make 102 grasp attempts on a set of test objects. The grasp attempts last for up to 20 time steps each, and any grasped object is deposited back into the bin. Although a policy may choose to grasp the same object multiple times, we found in practice that each robot made grasp attempts on a variety of objects, without fixating on a single one. However, to control for potential confounding effects due to replacement, we also conducted experiments with a second protocol, which we refer to as bin emptying. Here, a single robot unloads a cluttered bin filled with 28 test objects,
using 30 grasp attempts. This is repeated 5 times. Grasp success is reported over the first 10, 20, and 30 grasp attempts, corresponding to grasps on increasingly difficult objects.

\begin{table}[t]
\vspace{-0.05in}
{\footnotesize
\begin{center}
\begin{tabular}{ |p{7em}|p{14em}|p{3em}||p{3em}|p{3em}|p{3em}| } 
\hline
\multirow{2}{*}{Method} &
\multirow{2}{*}{Dataset} &
\multirow{2}{*}{Test} &
\multicolumn{3}{c|}{Bin emptying} \\
\cline{4-6}
&&& first 10 & first 20 & first 30 \\
\hline
QT-Opt (ours) & 580k off-policy $+$ 28k on-policy & {\bf 96\%} & {\bf 88\%} & {\bf 88\%} & {\bf 76\%} \\
 \hline
\citet{levine16} & 900k grasps from \citet{levine16} & 78\% & 76\% & 72\% & 72\% \\
\hline
QT-Opt (ours) & 580k off-policy grasps only & 87\% & \multicolumn{3}{}{} \\
\cline{1-3}
\citet{levine16} & 400k grasps from our dataset & 67\% & \multicolumn{3}{}{} \\
\cline{1-3}
\end{tabular}
\end{center}
}
\caption{Quantitative results in terms of grasp success rate on test objects. Policies are evaluated with object replacement (test) and without (bin emptying), with the latter showing success rates on the first 10, 20, and 30 grasps. The variant of our method that uses on-policy joint finetuning has a failure rate more than four times lower than prior work on the test set, while using substantially fewer grasp attempts for training. The variant that only uses off-policy training also substantially exceeds the performance of the prior method.
\label{tbl:quantitative_results}
\vspace{-0.3in}
}
\end{table}

The performance of our method is shown in Table~\ref{tbl:quantitative_results}. The results show both a variant of our method that is trained entirely using off-policy data, without any additional data collection from the latest policy, as well as the performance after joint finetuning with additional on-policy data, which is collected simultaneously with the policy training (details of the joint finetuning procedure in Appendix~\ref{sec:policy_fine_tuning}). The success rate of our method in both cases is very high. Effective off-policy training is valuable as it allows for rapid iteration on hyperparameters and architecture design without any data collection. However, additional on-policy joint finetuning consistently provides a quantifiable increase in performance with only about 28,000 additional grasps, reaching 96\% grasp success.
    Although the on-policy dataset does not observe the same data diversity as seen in the off-policy dataset, it likely affords the policy a kind of ``hard negative mining'' mechanism, letting it quickly correct erroneous and over-optimistic extrapolations. Further ablations are discussed in Appendix~\ref{sec:appendix_ablations_real}.

To compare our method to prior work, we evaluated the technique proposed by~\citet{levine16}. This prior method is also self-supervised, and previously attained good results on a similar visual grasping setup. This prior method does not reason about long-horizon rewards: although it can be used in closed-loop, the policy greedily optimizes for grasp success at the next grasp, does not control the opening and closing of the gripper, and does not reason about pregrasp manipulation. Since the format of the data for the two methods is different due to the different action representations, we compare to two versions of this prior approach: a variant that is trained on all of the data described by~\citet{levine16}, and a variant that adapts the same data used for our method, discarding grasp attempts where the gripper was not closed. The comparison in Table~\ref{tbl:quantitative_results} indicates a very large gap in performance between our method and both variants of the prior approach. On the bin emptying experiment, our method emptied the bin in 30 grasps or less in 2 of the 5 trials, while the prior method emptied the bin in 1 of the 5 trials. The lower success rate for 30 grasps is due to the policy trying to grasp the last few objects, which are usually very small and often get stuck in an unreachable corner of the bin. Examples are shown in Appendix~\ref{sec:appendix_ablations_real}.

\vspace{-0.1in}
\subsection{Analysis of Grasping Strategies with Qualitative Experiments}
\vspace{-0.05in}
\label{sec:qualitative}

\begin{figure}
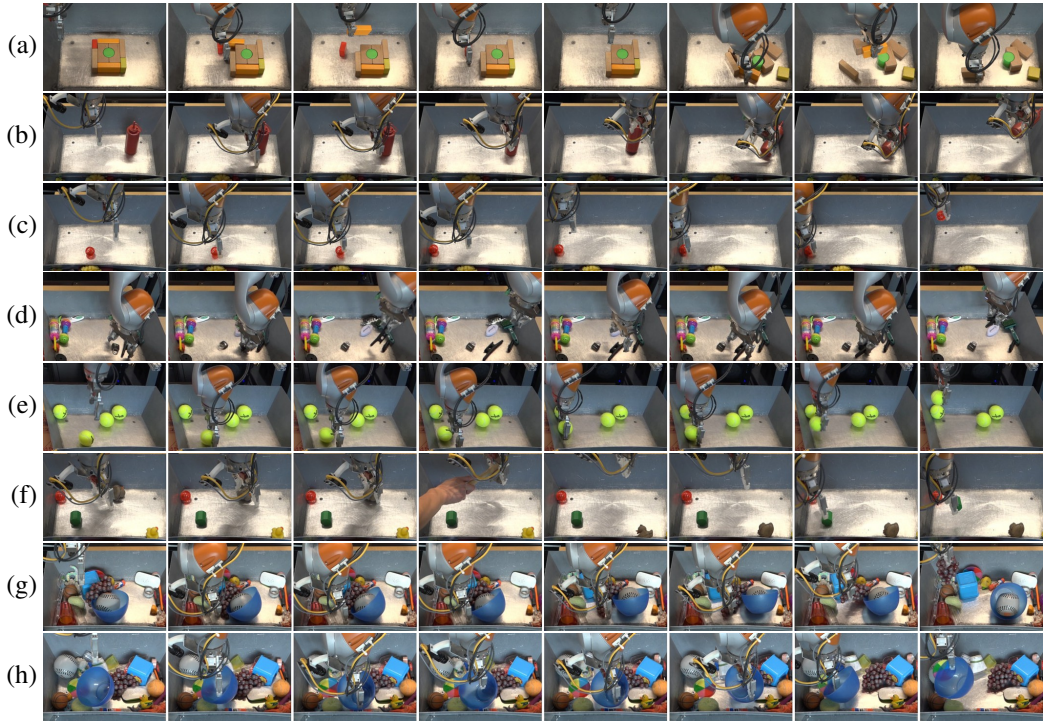

    \vspace{-0.05in}
    \centering
    \begin{imgrows}
        \imgrow \img{images/rows/image1}
        \imgrow \img{images/rows/image2}
        \imgrow \img{images/rows/image3}
        \imgrow \img{images/rows/image4}
        \imgrow \img{images/rows/image5}
        \imgrow \img{images/rows/image6}
        \imgrow \img{images/rows/image7}
        \imgrow \img{images/rows/image8}
    \end{imgrows}
    \caption{Eight grasps from the QT-Opt policy, illustrating some of the strategies discovered by our method: pregrasp manipulation (a, b), grasp readjustment (c, d), grasping dynamic objects and recovery from perturbations (e, f), and grasping in clutter (g, h). See discussion in the text and Appendix \ref{sec:appendix_ablations_real}.}
    \label{fig:qualitative}
    \vspace{-0.2in}
\end{figure}

Our QT-Opt grasping policy has a success rate of 96\% on previously unseen test objects. What types of strategies does this policy adopt? %
In contrast to most grasping systems, our method performs general closed-loop control with image observations, and can choose to reposition, open, or close the gripper at any time. This flexibility, combined with training for long-horizon success with reinforcement learning, enables it to perform behaviors that are usually not observed with standard grasping systems. %
We encourage the reader to watch the supplementary video, as well as the extended video, both provided at \siteurl, and discuss some examples here.
Notably, all of these examples emerge automatically from training the policy to optimize grasp success.

\vspace{-0.12in}
\paragraph{Singulation and pregrasp manipulation.}
Since our policies optimizes for the success of the entire episode, they can carry out pregrasp manipulations that reposition objects to make them easier to grasp. In Fig.~\ref{fig:qualitative} (a), we show an example object singulation sequence performed by the learned policy on a previously unseen blocks puzzle, and in Fig.~\ref{fig:qualitative} (b), we show an example where the policy chooses to knock down a ketchup bottle to make it easier to pick up.

\vspace{-0.12in}
\paragraph{Regrasping.}
The policy can open and close the gripper at any time, which allows it to detect early signs of an unstable grasp and regrasp the object more securely. In Fig.~\ref{fig:qualitative} (c), we show examples where the policy repeatedly regrasps a slippery object on the floor, while in Fig.~\ref{fig:qualitative} (d), we show an example where the object slips out of the gripper during the load phase, and the policy repositions the gripper for a more secure grasp.

\vspace{-0.12in}
\paragraph{Handling disturbances and dynamic objects.}
The reactive policy can also grasp objects that move dynamically during the grasping process. In Fig.~\ref{fig:qualitative} (e), we show examples where the policy attempts to pick up a ball, which rolls out of the gripper forcing the robot to follow. In Fig.~\ref{fig:qualitative} (f), we also show examples where the object is intentionally pushed out of the gripper during grasping. The policy is still able to correct and grasp another object successfully.

\vspace{-0.12in}
\paragraph{Grasping in clutter.}
Although the training data included no more than ten objects at a time, the policy can still grasp in dense clutter, as shown in Fig.~\ref{fig:qualitative} (g).

\vspace{-0.12in}
\paragraph{Failure cases.}
Although the policy was usually successful, we did observe a few failure cases. Especially in dense clutter, the policy was sometimes prone to regrasp repeatedly among cluttered objects, as shown in Fig.~\ref{fig:qualitative} (h). While this strategy often does produce a successful grasp, it is somewhat time consuming and not as goal-directed as the behavior observed in less cluttered scenes.

\vspace{-0.12in}
\section{Discussion and Future Work}
\label{sec:conclusion}

We presented a framework for scalable robotic reinforcement learning with raw sensory inputs such as images, based on an algorithm called QT-Opt, a distributed optimization framework, and a combination of off-policy and on-policy training. We apply this framework to the task of grasping, learning closed-loop vision-based policies that attain a high success rate on previously unseen objects, and exhibit sophisticated and intelligent closed-loop behavior, including singulation and pregrasp manipulation, regrasping, and dynamic responses to disturbances. All of these behaviors emerge automatically from optimizing the grasp success probability via QT-Opt. Although our policies are trained on a large amount of robot experience (\numgrasps{} real-world grasps), all of this experience is collected autonomously with minimal human intervention, and the amount of data needed is substantially lower than comparable prior self-supervised techniques (e.g., \citep{levine16}). Our results demonstrate that reinforcement learning with vision-based inputs can scale to large datasets and very large models, and can enable policies that generalize effectively for complex real-world tasks such as grasping. Our framework is generic with respect to the task, and extending the approach to other manipulation skills would be an exciting direction for future work.

\clearpage
\section*{Acknowledgments}
We would like to give special thanks to I{\~n}aki Gonzalo and John-Michael Burke for overseeing the robot operations, and Chelsea Finn, Timothy Lillicrap, and Arun Nair for valuable discussions.

\bibliography{references}

\begin{thebibliography}{47}
\providecommand{\natexlab}[1]{#1}
\providecommand{\url}[1]{\texttt{#1}}
\expandafter\ifx\csname urlstyle\endcsname\relax
  \providecommand{\doi}[1]{doi: #1}\else
  \providecommand{\doi}{doi: \begingroup \urlstyle{rm}\Url}\fi

\bibitem[Peters and Schaal(2008)]{peters_2008}
J.~Peters and S.~Schaal.
\newblock Reinforcement learning of motor skills with policy gradients.
\newblock \emph{Neural Networks}, 21\penalty0 (4):\penalty0 682 -- 697, 2008.
\newblock ISSN 0893-6080.
\newblock Robotics and Neuroscience.

\bibitem[Kalakrishnan et~al.(2011)Kalakrishnan, Righetti, Pastor, and
  Schaal]{kalakris11}
M.~Kalakrishnan, L.~Righetti, P.~Pastor, and S.~Schaal.
\newblock {Learning Force Control Policies for Compliant Manipulation}.
\newblock In \emph{IEEE/RSJ International Conference on Intelligent Robots and
  Systems}, 2011.

\bibitem[Yahya et~al.(2017)Yahya, Li, Kalakrishnan, Chebotar, and
  Levine]{yahya17}
A.~Yahya, A.~Li, M.~Kalakrishnan, Y.~Chebotar, and S.~Levine.
\newblock Collective robot reinforcement learning with distributed asynchronous
  guided policy search.
\newblock In \emph{IEEE/RSJ International Conference on Intelligent Robots and
  Systems}, 2017.

\bibitem[Ghadirzadeh et~al.(2017)Ghadirzadeh, Maki, Kragic, and
  Björkman]{kth17}
A.~Ghadirzadeh, A.~Maki, D.~Kragic, and M.~Björkman.
\newblock Deep predictive policy training using reinforcement learning.
\newblock In \emph{IEEE/RSJ International Conference on Intelligent Robots and
  Systems}, 2017.

\bibitem[Zeng et~al.(2018)Zeng, Song, Welker, Lee, Rodriguez, and
  Funkhouser]{zeng2018}
A.~Zeng, S.~Song, S.~Welker, J.~Lee, A.~Rodriguez, and T.~Funkhouser.
\newblock {Learning Synergies between Pushing and Grasping with Self-supervised
  Deep Reinforcement Learning}.
\newblock \emph{arXiv preprint arXiv:1803.09956}, 2018.

\bibitem[Morrison~{et. al}(2018)]{juxi18}
D.~Morrison~{et. al}.
\newblock {Cartman: The low-cost Cartesian Manipulator that won the Amazon
  Robotics Challenge}.
\newblock In \emph{IEEE International Conference on Robotics and Automation},
  2018.

\bibitem[Mahler et~al.(2017)Mahler, Matl, Liu, Li, Gealy, and
  Goldberg]{dexnet30_2017}
J.~Mahler, M.~Matl, X.~Liu, A.~Li, D.~V. Gealy, and K.~Goldberg.
\newblock {Dex-Net 3.0: Computing Robust Robot Suction Grasp Targets in Point
  Clouds using a New Analytic Model and Deep Learning}.
\newblock \emph{CoRR}, abs/1709.06670, 2017.
\newblock URL \url{http://arxiv.org/abs/1709.06670}.

\bibitem[ten Pas et~al.(2017)ten Pas, Gualtieri, Saenko, and
  Platt]{platt_gpd_17}
A.~ten Pas, M.~Gualtieri, K.~Saenko, and R.~Platt.
\newblock {Grasp Pose Detection in Point Clouds}.
\newblock \emph{The International Journal of Robotics Research}, 36\penalty0
  (13-14):\penalty0 1455--1473, 2017.

\bibitem[Chavan-Dafle and Rodriguez(2018)]{rodriguez2018icra}
N.~Chavan-Dafle and A.~Rodriguez.
\newblock {Stable Prehensile Pushing: In-Hand Manipulation with Alternating
  Sticking Contacts}.
\newblock In \emph{IEEE Intl Conference on Robotics and Automation}, 2018.

\bibitem[Bohg et~al.(2014)Bohg, Morales, Asfour, and Kragic]{bohg2014}
J.~Bohg, A.~Morales, T.~Asfour, and D.~Kragic.
\newblock {Data-Driven Grasp Synthesis — A Survey}.
\newblock \emph{IEEE Transactions on Robotics}, 30\penalty0 (2):\penalty0
  289--309, 2014.

\bibitem[Sutton and Barto(1998)]{sutton98}
R.~S. Sutton and A.~G. Barto.
\newblock \emph{{Introduction to Reinforcement Learning}}.
\newblock MIT Press, Cambridge, MA, USA, 1st edition, 1998.
\newblock ISBN 0262193981.

\bibitem[Tesauro(1994)]{tesauro94}
G.~Tesauro.
\newblock {TD-Gammon, a Self-Teaching Backgammon Program, Achieves Master-Level
  Play}.
\newblock \emph{Neural Computation}, March 1994.

\bibitem[Machado et~al.(2017)Machado, Bellemare, Talvitie, Veness, Hausknecht,
  and Bowling]{machado17arcade}
M.~C. Machado, M.~G. Bellemare, E.~Talvitie, J.~Veness, M.~J. Hausknecht, and
  M.~Bowling.
\newblock {Revisiting the Arcade Learning Environment: Evaluation Protocols and
  Open Problems for General Agents}.
\newblock \emph{CoRR}, abs/1709.06009, 2017.

\bibitem[Brockman et~al.(2016)Brockman, Cheung, Pettersson, Schneider,
  Schulman, Tang, and Zaremba]{gym16}
G.~Brockman, V.~Cheung, L.~Pettersson, J.~Schneider, J.~Schulman, J.~Tang, and
  W.~Zaremba.
\newblock Openai gym, 2016.

\bibitem[Hafner and Riedmiller(2011)]{hafner11}
R.~Hafner and M.~Riedmiller.
\newblock Reinforcement learning in feedback control.
\newblock \emph{Machine Learning}, 84\penalty0 (1-2), 2011.

\bibitem[Lillicrap et~al.(2015)Lillicrap, Hunt, Pritzel, Heess, Erez, Tassa,
  Silver, and Wierstra]{lillicrap15}
T.~P. Lillicrap, J.~J. Hunt, A.~Pritzel, N.~Heess, T.~Erez, Y.~Tassa,
  D.~Silver, and D.~Wierstra.
\newblock {Continuous Control with Deep Reinforcement Learning}.
\newblock \emph{CoRR}, abs/1509.02971, 2015.
\newblock URL \url{http://arxiv.org/abs/1509.02971}.

\bibitem[Duan et~al.(2016)Duan, Chen, Houthooft, Schulman, and Abbeel]{duan16}
Y.~Duan, X.~Chen, R.~Houthooft, J.~Schulman, and P.~Abbeel.
\newblock {Benchmarking Deep Reinforcement Learning for Continuous Control}.
\newblock In \emph{Intl Conference on Machine Learning}, 2016.

\bibitem[Henderson et~al.(2017)Henderson, Islam, Bachman, Pineau, Precup, and
  Meger]{deeprlthatmatters17}
P.~Henderson, R.~Islam, P.~Bachman, J.~Pineau, D.~Precup, and D.~Meger.
\newblock {Deep Reinforcement Learning that Matters}.
\newblock \emph{CoRR}, 2017.
\newblock URL \url{http://arxiv.org/abs/1709.06560}.

\bibitem[Kahn et~al.(2018)Kahn, Villaflor, Ding, Abbeel, and Levine]{gcg}
G.~Kahn, A.~Villaflor, B.~Ding, P.~Abbeel, and S.~Levine.
\newblock {Self-Supervised Deep Reinforcement Learning with Generalized
  Computation Graphs for Robot Navigation}.
\newblock In \emph{IEEE International Conference on Robotics and Automation},
  2018.

\bibitem[Quillen et~al.(2018)Quillen, Jang, Nachum, Finn, Ibarz, and
  Levine]{quillen}
D.~Quillen, E.~Jang, O.~Nachum, C.~Finn, J.~Ibarz, and S.~Levine.
\newblock {Deep Reinforcement Learning for Vision-Based Robotic Grasping: A
  Simulated Comparative Evaluation of Off-Policy Methods}.
\newblock In \emph{IEEE International Conference on Robotics and Automation},
  2018.

\bibitem[Levine et~al.(2016)Levine, Finn, Darrell, and Abbeel]{levine2015}
S.~Levine, C.~Finn, T.~Darrell, and P.~Abbeel.
\newblock {End-to-end Training of Deep Visuomotor Policies}.
\newblock \emph{Journal of Machine Learning Research}, 17\penalty0 (39), 2016.

\bibitem[Weisz and Allen(2012)]{weisz2012}
J.~Weisz and P.~K. Allen.
\newblock {Pose Error Robust Grasping from Contact Wrench Space Metrics}.
\newblock In \emph{IEEE International Conference on Robotics and Automation},
  2012.

\bibitem[Lenz et~al.(2015)Lenz, Lee, and Saxena]{lenz2015}
I.~Lenz, H.~Lee, and A.~Saxena.
\newblock {Deep Learning for Detecting Robotic Grasps}.
\newblock \emph{The International Journal of Robotics Research}, 34\penalty0
  (4-5):\penalty0 705--724, 2015.

\bibitem[Yu and Rodriguez(2018)]{yu2018icra}
K.~Yu and A.~Rodriguez.
\newblock {Realtime State Estimation with Tactile and Visual sensing.
  Application to Planar Manipulation}.
\newblock In \emph{IEEE Intl Conference on Robotics and Automation}, 2018.

\bibitem[Viereck et~al.(2017)Viereck, ten Pas, Saenko, and Platt]{platt17}
U.~Viereck, A.~ten Pas, K.~Saenko, and R.~Platt.
\newblock {Learning a visuomotor controller for real world robotic grasping
  using simulated depth images}.
\newblock In \emph{CoRL}, 2017.

\bibitem[Hausman et~al.(2017)Hausman, Chebotar, Kroemer, Sukhatme, and
  Schaal]{hausman17}
K.~Hausman, Y.~Chebotar, O.~Kroemer, G.~S. Sukhatme, and S.~Schaal.
\newblock {Regrasping using Tactile Perception and Supervised Policy Learning}.
\newblock In \emph{AAAI Symposium on Interactive Multi-Sensory Object
  Perception for Embodied Agents}, 2017.

\bibitem[Levine et~al.(2016)Levine, Pastor, Krizhevsky, and Quillen]{levine16}
S.~Levine, P.~Pastor, A.~Krizhevsky, and D.~Quillen.
\newblock Learning hand-eye coordination for robotic grasping with large-scale
  data collection.
\newblock In \emph{International Symposium on Experimental Robotics}, 2016.

\bibitem[Pinto and Gupta(2016)]{pinto16}
L.~Pinto and A.~Gupta.
\newblock {Supersizing self-supervision: Learning to grasp from 50K tries and
  700 robot hours}.
\newblock In \emph{IEEE International Conference on Robotics and Automation},
  2016.

\bibitem[Morrison et~al.(2018)Morrison, Corke, and Leitner]{morrison18}
D.~Morrison, P.~Corke, and J.~Leitner.
\newblock {Closing the Loop for Robotic Grasping: {A} Real-time, Generative
  Grasp Synthesis Approach}.
\newblock In \emph{Robotics: Science and Systems}, 2018.

\bibitem[Mnih et~al.(2015)]{mnih2015}
V.~Mnih et~al.
\newblock {Human-level control through deep reinforcement learning}.
\newblock \emph{Nature}, 518\penalty0 (7540):\penalty0 529--533, 2015.

\bibitem[Gu et~al.(2016)Gu, Lillicrap, Sutskever, and Levine]{gu16}
S.~Gu, T.~Lillicrap, I.~Sutskever, and S.~Levine.
\newblock {Continuous Deep Q-learning with Model-based Acceleration}.
\newblock In \emph{Proceedings of Intl Conference on Machine Learning}, 2016.

\bibitem[Polyak and Juditsky(1992)]{polyak1992acceleration}
B.~T. Polyak and A.~B. Juditsky.
\newblock Acceleration of stochastic approximation by averaging.
\newblock \emph{SIAM Journal on Control and Optimization}, 30\penalty0
  (4):\penalty0 838--855, 1992.

\bibitem[Lillicrap et~al.(2016)Lillicrap, Hunt, Pritzel, Heess, Erez, Tassa,
  Silver, and Wierstra]{lhph-ccdrl-16}
T.~Lillicrap, J.~Hunt, A.~Pritzel, N.~Heess, T.~Erez, Y.~Tassa, D.~Silver, and
  D.~Wierstra.
\newblock Continuous control with deep reinforcement learning.
\newblock In \emph{International Conference on Learning Representations}, 2016.

\bibitem[Hasselt(2010)]{hasselt10}
H.~V. Hasselt.
\newblock {Double Q-learning}.
\newblock In J.~D. Lafferty, C.~K.~I. Williams, J.~Shawe-Taylor, R.~S. Zemel,
  and A.~Culotta, editors, \emph{Advances in Neural Information Processing
  Systems}, 2010.

\bibitem[Hasselt et~al.(2016)Hasselt, Guez, and Silver]{hasselt16}
H.~v. Hasselt, A.~Guez, and D.~Silver.
\newblock {Deep Reinforcement Learning with Double Q-Learning}.
\newblock In \emph{AAAI Conference on Artificial Intelligence}, 2016.

\bibitem[Fujimoto et~al.(2018)Fujimoto, van Hoof, and Meger]{td3}
S.~Fujimoto, H.~van Hoof, and D.~Meger.
\newblock {Addressing Function Approximation Error in Actor-Critic Methods}.
\newblock \emph{CoRR}, 2018.
\newblock URL \url{http://arxiv.org/abs/1802.09477}.

\bibitem[Amos et~al.(2017)Amos, Xu, and Kolter]{amos2017icnn}
B.~Amos, L.~Xu, and J.~Z. Kolter.
\newblock Input convex neural networks.
\newblock In \emph{International Conference on Machine Learning}, volume~70,
  pages 146--155, 2017.

\bibitem[Rubinstein and Kroese(2004)]{rk-cem-04}
R.~Rubinstein and D.~Kroese.
\newblock \emph{{The Cross-Entropy Method}}.
\newblock Springer-Verlag, 2004.

\bibitem[Coumans and Bai(2016--2018)]{bullet}
E.~Coumans and Y.~Bai.
\newblock Pybullet, a python module for physics simulation for games, robotics
  and machine learning.
\newblock \url{http://pybullet.org}, 2016--2018.

\bibitem[Chang et~al.(2015)Chang, Funkhouser, Guibas, Hanrahan, Huang, Li,
  Savarese, Savva, Song, Su, Xiao, Yi, and Yu]{shapenet}
A.~X. Chang, T.~A. Funkhouser, L.~J. Guibas, P.~Hanrahan, Q.~Huang, Z.~Li,
  S.~Savarese, M.~Savva, S.~Song, H.~Su, J.~Xiao, L.~Yi, and F.~Yu.
\newblock Shapenet: An information-rich 3d model repository.
\newblock \emph{CoRR}, abs/1512.03012, 2015.
\newblock URL \url{http://arxiv.org/abs/1512.03012}.

\bibitem[Ioffe and Szegedy(2015)]{is-bnad-15}
S.~Ioffe and C.~Szegedy.
\newblock {Batch Normalization: Accelerating Deep Network Training by Reducing
  Internal Covariate Shift}.
\newblock In \emph{International Conference on Machine Learning}, 2015.

\bibitem[Stooke and Abbeel(2018)]{stooke_accelerated}
A.~Stooke and P.~Abbeel.
\newblock {Accelerated Methods for Deep Reinforcement Learning}.
\newblock \emph{CoRR}, abs/1803.02811, 2018.
\newblock URL \url{http://arxiv.org/abs/1803.02811}.

\bibitem[Espeholt et~al.(2018)Espeholt, Soyer, Munos, Simonyan, Mnih, Ward,
  Doron, Firoiu, Harley, Dunning, Legg, and Kavukcuoglu]{impala}
L.~Espeholt, H.~Soyer, R.~Munos, K.~Simonyan, V.~Mnih, T.~Ward, Y.~Doron,
  V.~Firoiu, T.~Harley, I.~Dunning, S.~Legg, and K.~Kavukcuoglu.
\newblock {IMPALA:} scalable distributed deep-rl with importance weighted
  actor-learner architectures.
\newblock \emph{CoRR}, abs/1802.01561, 2018.
\newblock URL \url{http://arxiv.org/abs/1802.01561}.

\bibitem[Nair et~al.(2015)Nair, Srinivasan, Blackwell, Alcicek, Fearon, Maria,
  Panneershelvam, Suleyman, Beattie, Petersen, Legg, Mnih, Kavukcuoglu, and
  Silver]{gorila}
A.~Nair, P.~Srinivasan, S.~Blackwell, C.~Alcicek, R.~Fearon, A.~D. Maria,
  V.~Panneershelvam, M.~Suleyman, C.~Beattie, S.~Petersen, S.~Legg, V.~Mnih,
  K.~Kavukcuoglu, and D.~Silver.
\newblock Massively parallel methods for deep reinforcement learning.
\newblock \emph{CoRR}, abs/1507.04296, 2015.
\newblock URL \url{http://arxiv.org/abs/1507.04296}.

\bibitem[Horgan et~al.(2018)Horgan, Quan, Budden, Barth-Maron, Hessel, van
  Hasselt, and Silver]{horgan2018distributed}
D.~Horgan, J.~Quan, D.~Budden, G.~Barth-Maron, M.~Hessel, H.~van Hasselt, and
  D.~Silver.
\newblock {Distributed Prioritized Experience Replay}.
\newblock In \emph{International Conference on Learning Representations}, 2018.

\bibitem[Dean et~al.(2012)Dean, Corrado, Monga, Chen, Devin, Le, Mao, Ranzato,
  Senior, Tucker, Yang, and Ng]{dean_async}
J.~Dean, G.~S. Corrado, R.~Monga, K.~Chen, M.~Devin, Q.~V. Le, M.~Z. Mao,
  M.~Ranzato, A.~Senior, P.~Tucker, K.~Yang, and A.~Y. Ng.
\newblock Large scale distributed deep networks.
\newblock In \emph{Advances in Neural Information Processing Systems}, 2012.

\bibitem[Abadi~{et. al}(2015)]{tensorflow2015-whitepaper}
M.~Abadi~{et. al}.
\newblock {TensorFlow}: Large-scale machine learning on heterogeneous systems,
  2015.
\newblock URL \url{https://www.tensorflow.org/}.

\end{thebibliography}

\newpage

\appendix

\section{Real World Ablation Experiments: State, Action, and Reward Design}
\label{sec:appendix_ablations_real}
After prototyping ideas in our simulated setup, we take the best parameters discussed in Appendix~\ref{sec:appendix_ablations_sim}, and repeat those experiments in the real setup, to verify the same parameters carry over across domains. Since real-world finetuning takes considerable robot time, all of these experiments were conducted with entirely off-policy training using a fixed dataset of 580k grasps. The results are therefore in absolute terms worse than the final results of our best policy, but are still useful for understanding the relative tradeoffs of various design choices.

\vspace{-0.1in}
\paragraph{State representation and its effects.}

Echoing results discussed in Appendix~\ref{sec:appendix_ablations_sim} we found that a rich state representation greatly impacts real robot performance. Matching simulated experiments, providing the image, gripper status, and height to bottom of the bin performs better than other representations. These experiments indicate that while hand-eye coordination can in principle be figured out purely from the image observation, explicitly adding domain-specific state features improves performance. We find it very important that such low dimensional state features can be seamlessly integrated into our model, resulting in better performance and data efficiency. All models are trained off-policy for 2.5M steps, with discount $0.9$ and no reward penalty.

\begin{table}[h]
\begin{center}
\begin{tabular}{ |p{12em}|p{7em}| } 
\hline
State Representation & Performance \\
\hline
Image only & 53\% \\ %
\hline
Image + gripper status & 58\% \\ %
\hline
Image + gripper status + height & 70\% \\ %
\hline
\end{tabular}
\end{center}
\caption{Off-policy ablation over state representation.}
\label{table:state-repr-ablation}
\vspace{-0.25in}
\end{table}
\paragraph{Discount and Reward Definition}
To encourage faster grasps, we experimented with decreasing discount and adding a small reward penalty at each timestep. Again, matching sim results, a reward penalty did better than decreasing the discount factor. All models are trained off-policy on the same dataset for 2.5M steps.
\begin{table}[h]
\label{table:learning-ablation}
\begin{center}
\begin{tabular}{|p{12em}|p{7em}|p{7em}|p{7em}|}
\hline
State Representation & Discount Factor & Reward Penalty & Performance  \\
\hline
Image only & 0.9 & 0 & 53\% \\ %
\hline
Image only & 0.7 & 0 & 28\% \\ %
\hline
Image only & 0.9 & -0.05 & 63\% \\ %
\hline
\end{tabular}
\end{center}
\caption{Off-policy ablation over discount and reward.}
\vspace{-0.3in}
\end{table}
\paragraph{Learned Termination}
We compare a task-specific scripted termination condition with a task-agnostic termination action learned by the policy. Details of the scripted termination and learned termination conditions are in the Appendix \ref{appendix:termination}. The learned termination condition performs better in the off-policy case and on-policy case.
 
\begin{table}[h]
\label{table:termination-ablation}
\begin{center}
\begin{tabular}{|p{12em}|p{9em}|p{7em}|}
\hline
Termination Condition & Training Regime & Performance  \\
\hline
Automatic & off-policy & 81\% \\ 
\hline
Learned & off-policy & 87\% \\ 
\hline
Automatic & on-policy joint finetuning & 95\% \\ 
\hline
Learned & on-policy joint finetuning & 96\% \\ 
\hline
\end{tabular}
\end{center}
\caption{Off-policy and on-policy ablation of termination condition.}
\vspace{-0.25in}
\end{table}
\paragraph{Quantitative experiments}
\begin{figure}[h]
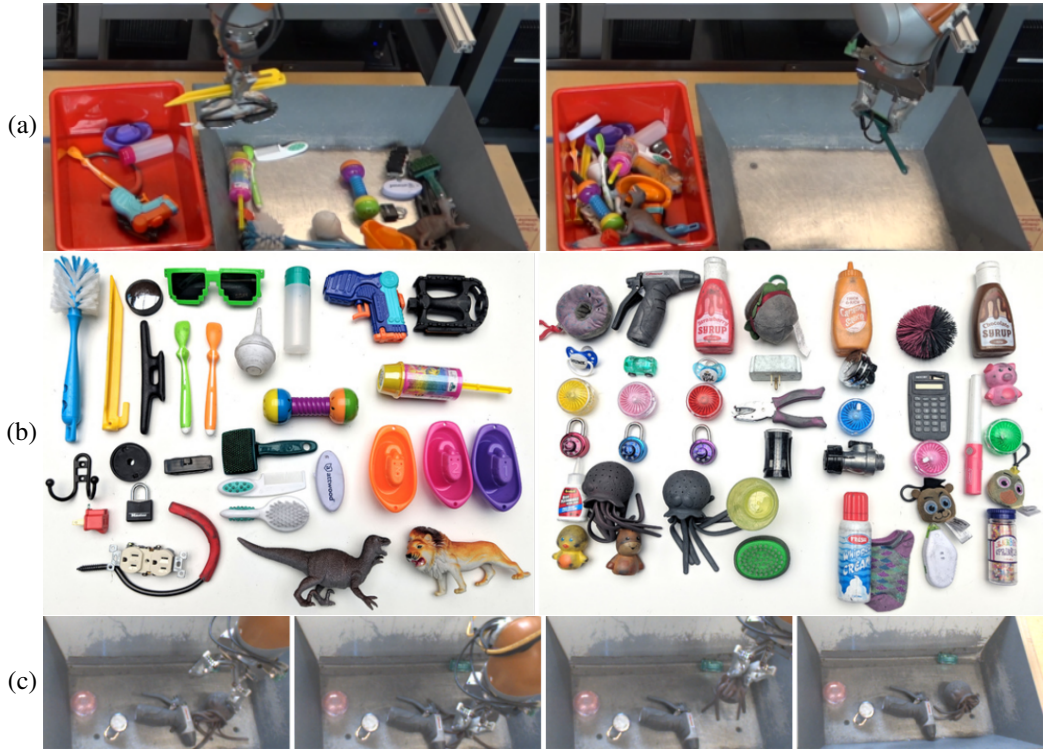

    \vspace{-0.05in}
    \centering
    \begin{imgrows}
        \imgrow \img{images/bin-unloading}
        \imgrow \img{images/test_two_protocols_side_by_side}
        \imgrow \img{images/octopus}
    \end{imgrows}
    \caption{Illustrations of the bin emptying experiment (a). The (a, right) shows a very small object getting stuck in the corner and requiring a few attempts to get grasped. The test objects for the bin emptying experiment (b, left) and grasping with replacement experiment (b, right). Note that the objects range greatly in size and appearance, and many are exceedingly difficult to grasp due to small size, complex shapes, and varied material properties. Sequence (c) illustrates a grasp on the legs of a toy octopus, %
    getting positive visual feedback and resulting in episode termination by the policy. During further ascent, the octopus slips out of the gripper, and the grasp is labeled as a failure. This is not a principled limitation of our method, as we could have terminated an episode at a larger height to get more robust feedback. Robot's observation $I_t$ is shown in (c).}
    \label{fig:bin_unloading}
    \vspace{-0.2in}
\end{figure}
The performance of our algorithm is evaluated empirically in a set of grasping experiments. As discussed in Section~\ref{sec:quantitative} we follow two experimentation protocols. In one set of the experiments we let 7 robots execute 102 grasps each and then average grasp success. See results in column three of Table~\ref{tbl:quantitative_results}. After a successful trial, the object is randomly dropped back into the bin. In a second protocol we place 28 objects into a bin and let the robot perform 30 grasps without object replacement. This time successfully grasped objects are placed into a spare basket. For the bin unloading protocol we report success rates on the first 10 / 20 / 30 grasps to report on increasingly difficult objects remaining in the bin over the unloading process. As the unloading process progresses, large objects tend to be grasped first and small objects are grasped last. The policy is required to localize the small objects and then make a grasp. The ability of our policy to localize small objects in a very sparse scene is learned solely from data, and was not hand tuned.

The test sets, shown in Fig.~\ref{fig:bin_unloading} (b), consist of objects that pose a variety of challenges for grasping. They range in size from very small to larger, heavier objects. They vary in appearance, and include objects that are translucent and reflective. They include objects with highly non-convex shapes, soft and deformable parts, and parts that are inherently unsuitable for grasping (such as the bristles on a brush).
In analyzing the failure cases, we noticed the following patterns. In the test with replacement, many of the failed grasps were on the soft octopus toy (Fig.~\ref{fig:bin_unloading} (c)), where the robot would lift the octopus successfully by one of the tentacles, but it would slip out and fall back into the bin after lifting. Other failures were caused by the small round objects rolling into the corners of the bin, where they were difficult to reach. Since we impose hard bounds on the workspace to prevent the robot from colliding with the environment, in many of these cases it was actually impossible for the robot to grasp the object successfully, regardless of the policy.
In the bin emptying experiment, we found that many of the failures were due to the small black lock (Fig.~\ref{fig:bin_unloading} (a, right)) getting stuck in the corner of the bin, where it could not be grasped, resulting in multiple sequential failed attempts.

\paragraph{Emergent grasping behaviors}
We presented a challenging task of grasping a toy puzzle which needs to be broken down into pieces before any individual part could be grasped, as otherwise the puzzle will not fit into the gripper, see Fig.~\ref{fig:qualitative} (a). This puzzle was not seen at training time. We treat grasp success as a proxy for frequency and efficiency of a pregrasp manipulation, since the first block cannot be grasped a without pregrasp manipulation. After every grasp, we reassemble the puzzle and place it at a different location. We compare our best baseline based on~\citet{levine16} to QT-Opt. The QT-Opt model succeeds in 19 out of 24 grasps (79\%), while the prior method succeeds in 13 out of 24 grasps (54\%).

In the second evaluation, we attempt grasps on a tennis ball, see Fig.~\ref{fig:qualitative} (e). The first time the tennis ball is between the gripper fingers, we push the tennis ball out of the way. The ball cannot be grasped unless the model reacts to the new tennis ball position. The QT-Opt model succeeds in 28 of 38 grasps (74\%). The prior model succeeds in 4 out of 25 grasps (16\%).

This suggests that our policy learned better proactive and reactive behaviors than our baseline.
\paragraph{Clipped Double Q-Learning}
We find that using Clipped Double Q-Learning~\cite{td3} results in faster convergence in simulated experiments (Figure~\ref{fig:sim-dqn-graphs}) and is crucial for good performance in real experiments (Table~\ref{table:real-qlearn-ablation}). Experiments used the scripted termination. 
\begin{table}[h]
\begin{center}
\begin{tabular}{|p{13em}|p{5em}|}
\hline
Q-learning Method & Performance  \\
\hline
Double Q-learning & 63\% \\ %
\hline
Clipped Double Q-learning & 81\% \\ 
\hline
\end{tabular}
\end{center}
\caption{Off-policy performance with and without clipped Double-Q Learning.}
\label{table:real-qlearn-ablation}
\vspace{-0.25in}
\end{table}
\paragraph{Data efficiency}
As discussed in Section~\ref{sec:grasping_system} we collected 580k grasp attempts across 7 robots with a total of about 800 robot hours. Since we saved collected data on disk, we can study performance of a policy with less off-policy data. A dataset of 320k grasp attempts was generated by using data from the first 440 robot hours of data collection.
Table~\ref{tbl:real-data-efficiency} shows performance of off-policy training, using the best model configuration. With only 55\% of the original dataset, we reached 78\% grasp success, the same performance as our best supervised learning baseline, but using one third the number of grasps, and half the number of transitions. Further joint finetuning would likely yield a final policy that also reaches 96\% grasp success, with higher data efficiency but more on-robot joint finetuning time.
\begin{table}[h]
\begin{center}
\begin{tabular}{|p{7em}|p{5em}|}
\hline
Dataset Size & Performance  \\
\hline
580k grasps & 87\% \\ 
\hline
320k grasps & 78\% \\ 
\hline
\end{tabular}
\end{center}
\caption{Data efficiency.}
\label{tbl:real-data-efficiency}
\vspace{-0.25in}
\end{table}

\section{Exploration and Dataset Bootstrapping}
\label{sec:appendix_exploration}
As is standard in Q-learning, we evaluate using one policy (evaluation policy) and collect training data with a different policy (exploration policy). Our evaluation policy $\pi_{eval}$ chooses each action by maximizing the Q-function value using our QT-Opt algorithm described in Section~\ref{sec:qtopt-details}. For data collection, we used two different exploration policies $\pi_{scripted}, \pi_{noisy}$ at different stages of training.

During the early stages of training, a policy that takes random actions would achieve reward too rarely to learn from, since the grasping task is a multi-stage problem and reward is sparse. This was indeed what we observed during early experimentation. For this reason, we collect our initial data for training using a scripted policy
\(\pi_{scripted}\) that successfully grasps 15-30\% of the time. The \(\pi_{scripted}\) simplifies the multi-step exploration of the problem by randomly choosing an $(x,y)$ coordinate above the table, lowering the open gripper to table level in a few random descent steps, closing the gripper, then returning to the original height in a few ascent steps.

We compared initial data collection with \(\pi_{scripted}\) vs. initial data collection with \(\pi_{prior}\), a grasping model based on~\citet{levine16}. In our real-world experiments, we used \(\pi_{prior}\), but in a simulated comparison, we found that initializing training with either policy leads to the same final performance and data efficiency. The data generated by either policy has similar distributional properties, as discussed in Appendix~\ref{sec:appendix_off_policy}, which is sufficient to bootstrap learning.

During the later stages of training, we switch to data collection with \(\pi_{noisy}\). This exploration policy uses epsilon-greedy exploration to trade off between choosing exploration actions or actions that maximize the Q-function estimate. The policy \(\pi_{noisy}\) chooses a random action with probability $\epsilon=20\%$, otherwise the greedy action is chosen. To choose a random action, \(\pi_{noisy}\) samples a pose change $\mathbf{t}, \mathbf{r}$ from a Gaussian with probability 75\%, a toggle gripper action $\mathbf{g_\text{open}}, \mathbf{g_\text{close}}$ with probability 17\%, and an episode termination $\mathbf{e}$ with probability 8\%.

\section{Simulated Experiments: Dataset Size, Off-Policy Training, MDP Design}
\label{sec:appendix_ablations_sim}
\begin{wrapfigure}{r}{0.22\textwidth}
\vspace{-0.4cm}
 \includegraphics[width=0.22\textwidth]{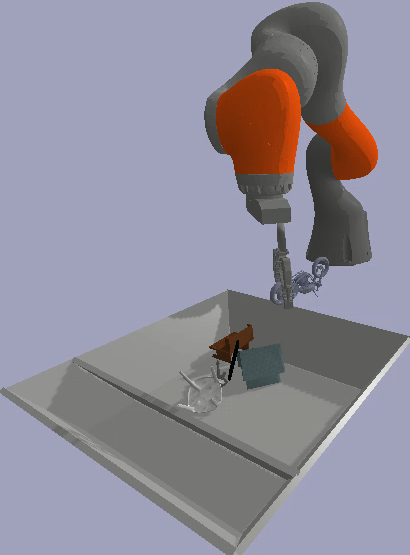}
\vspace{-0.4cm}
   \caption{Bullet Physics simulated environment.}
\vspace{-0.5cm}
\end{wrapfigure}
Robot experiments at large scale are excessively more time-consuming, complex, and uncertain than experimentation in a simulated environment.
We made extensive use of simulation for prototyping and ablation studies of our methods.

Our simulation environment is based on the Bullet Physics simulator~\citep{bullet} and mimics the setup shown in Fig.~\ref{fig:robot_setup_and_objects}. The simulated environment uses the configuration of the real Kuka IIWA arm, a similar looking bin, and a simulated over-the-shoulder camera. We used object models from the ShapeNet dataset ~\citep{shapenet}, scaled down to graspable sizes. Using our scalable distributed learning infrastructure (as discussed in Section~\ref{sec:qtopt}), we were able to generate data with up to 1,000 virtual robots running in parallel, and conduct a large scale experiment within a few hours. Both simulation and real used the same input modality, neural net architecture, and method of robotic control. Simulation was only used for prototyping, and all real world policies used only real data. We found that real world learning was generally much harder in terms of data requirements and time needed to train a good model, due to higher visual diversity, real world physics, and unmodeled properties of the real robot.

\begin{figure}[h]
    \centering
    \begin{subfigure}[t]{0.45\textwidth}
        \centering
        \includegraphics[width=\textwidth]{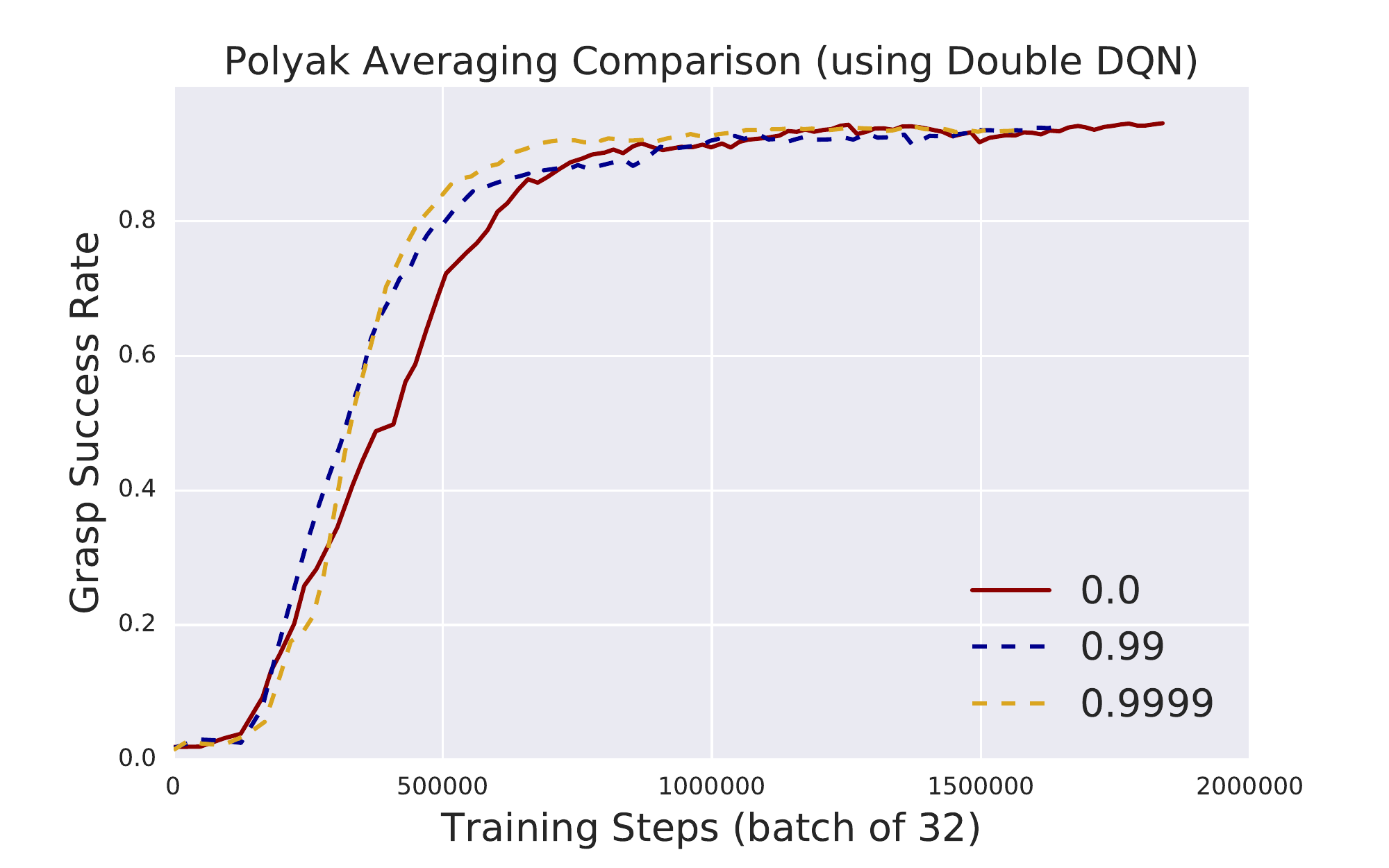}
        \caption{All values of Polyak averaging provide stable training that reach similar performance, but higher values seem to converge faster and have smoother performance curves.}
        \label{fig:polyak-comparison}
    \end{subfigure}
    ~
    \begin{subfigure}[t]{0.45\textwidth}
        \centering
        \includegraphics[width=\textwidth]{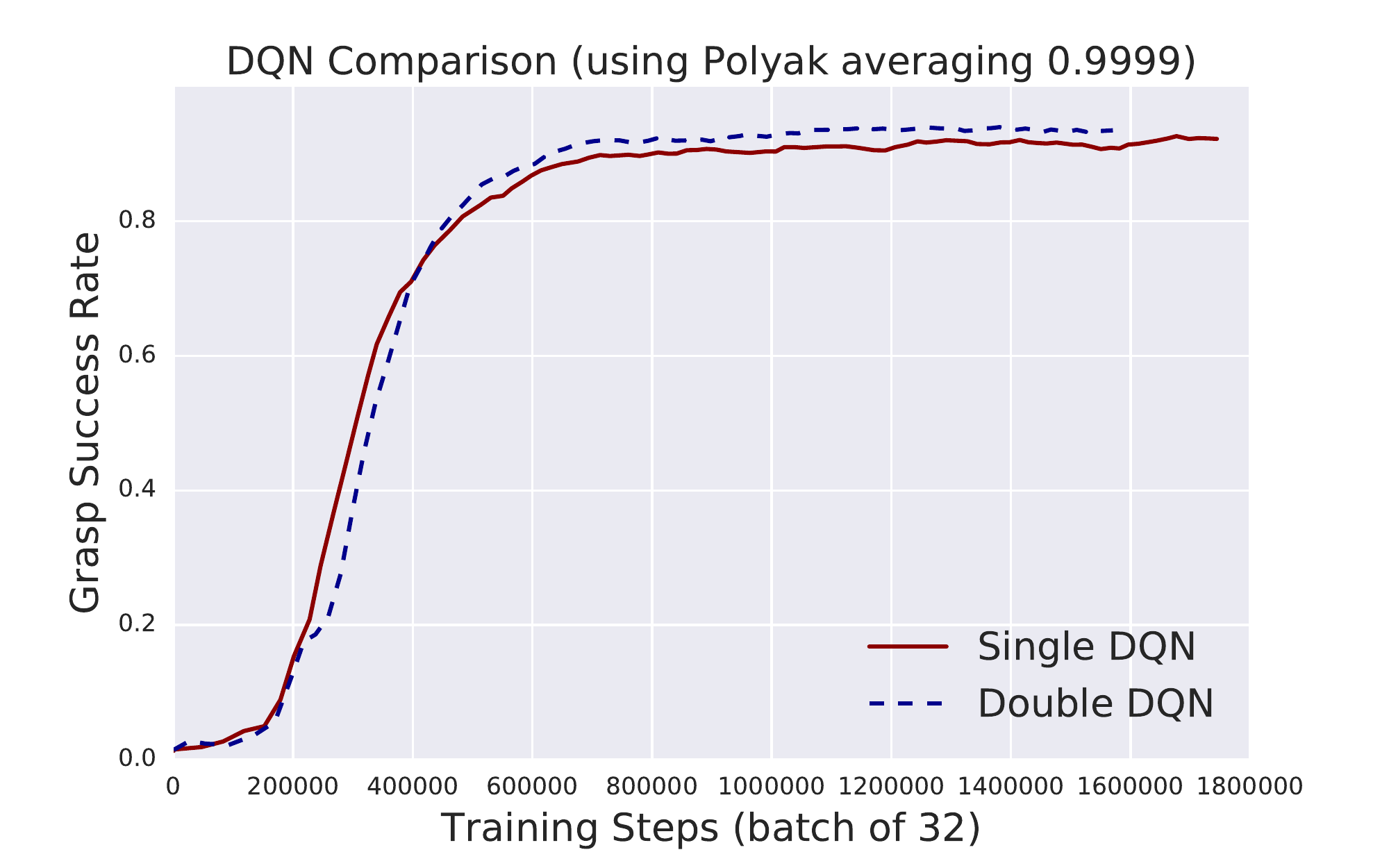}
        \caption{Double DQN outperforms Single DQN on our task. Note that both are quite stable.}
        \label{fig:dqn-comparison}
    \end{subfigure}
    \caption{Comparison of Polyak averaging constants (a) and Single DQN vs Double DQN (b). Note these experiments do not include clipped Double Q-learning.}
    \label{fig:qtopt-comparison}
\end{figure}

There are many parameters in our system which impact the final performance, data efficiency, generalization and emergence of rich behaviours. We split the factors into three large groups: QT-Opt specific parameters, grasping parameters, and data efficiency, which we discuss below.
\vspace{-0.1in}
\paragraph{QT-Opt specific parameters} include hyperparameters for Polyak averaging and the method of computing Bellman error. We found that $0.9999$ was the best Polyak averaging constant, and Double DQN performed better than Single DQN. Results are in Figure~\ref{fig:qtopt-comparison}. Note that although asymptotic performance is similar, the settings chosen seem to provide faster convergence and lower variance in grasp success. The latter is important for real experiments, where it is costly to evaluate the policy at multiple points. We also found that the cross-entropy loss performed better than the standard squared difference loss.

\begin{figure}[h]
\centering
 \includegraphics[width=0.5\textwidth]{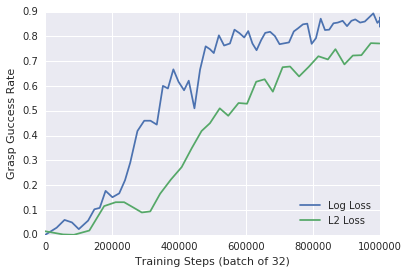}
   \caption{Loss function comparison in simulation}
\label{fig:sim-loss-graphs}
\end{figure}

\begin{figure}[h]
\centering
 \includegraphics[width=0.5\textwidth]{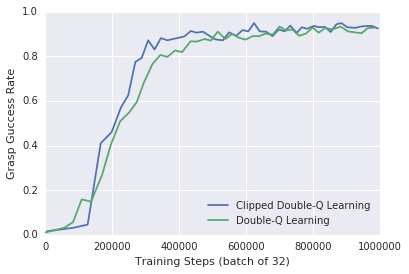}
   \caption{Comparison of DQN methods. See Appendix~\ref{sec:appendix_bellman_update} for clipped Double DQN definition.}
\label{fig:sim-dqn-graphs}
\vspace{-0.1in}
\end{figure}

Figure~\ref{fig:sim-dqn-graphs} compares Clipped Double DQN to standard Double DQN, showing Clipped Double DQN does slightly better. Note that although the difference in sim was small, we found performance was significantly better in the real world off-policy setting (see Table~\ref{table:real-qlearn-ablation}). The simulated setup uses 60 simulated, on-policy robots, which suggests the primary gains of Clipped Double DQN come when it is used with off-policy data.

\paragraph{Grasping parameters} describe the task specific MDP design, \ie state-action representations, closed-loop control and data generation policies. We found that differences in state representation as discussed in detail in Appendix~\ref{sec:appendix_state_action_reward}, greatly impacts performance. We compared models whose state is either a) just an image observation of the scene or b) a richer state comprising an image, the gripper aperture and distance between the gripper and the floor. Next, we studied discount factors and reward penalties. Finally, we compared a scripted episode stopping criterion to a learned termination action. In these experiments, we use 60 simulated robots, using \(\pi_{scripted}\) policy for the first 120K gradient update steps, then switching to the \(\pi_{noisy}\) policy with \(\epsilon=0.2\). These exploration policies are explained in Appendix~\ref{sec:appendix_exploration}. The grasp performance is evaluated continuously and concurrently as training proceeds by running the \(\pi_{eval}\) policy on 100 separate simulated robots and aggregating 700 grasps per policy for each model checkpoint.

\begin{table}[h]
\begin{center}
\begin{tabular}{ |p{7em}|p{5em}|p{5em}|p{5em}|p{5em}|p{5em}| } 
\hline
State & Termination action & Intermediate reward & Discount factor & Perf. at 300K steps & Perf. at 1M steps\\
\hline
\multirow{3}{7em}{Image+gripper status+height}
& No & -0.05 & 0.9 & 75\%& 95\% \\ %
& No & 0 & 0.9 & 68\%& 92\% \\ %
& No & 0 & 0.7 & 50\%& 90\% \\ %
\hline
Image only & No & -0.05 & 0.9 & 25\%& 81\% \\ %
\hline
Image+gripper status+height & Yes & -0.05 & 0.9 & 67\%& 94\% \\
\hline
\end{tabular}
\end{center}
\caption{Simulation studies for tuning grasping task parameters}
\label{table:sim-grasping-ablation}
\end{table}

\begin{figure}[h]
\centering
 \includegraphics[width=0.5\textwidth]{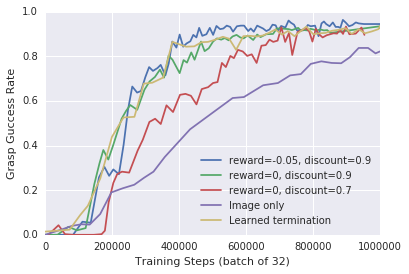}
   \caption{Performance graphs of simulation studies for tuning grasping task parameters}
\label{fig:sim-graphs}
\vspace{-0.05in}
\end{figure}
The results in Table~\ref{table:sim-grasping-ablation} show that richer state representation results in faster convergence and better final performance. A small reward penalty does much better than decreasing the discount factor. Finally, giving the policy greater control over the episode termination yields performance on par with the engineered termination condition.

We put special focus on the termination action: by letting the policy decide when to terminate, the policy has more potential to keep trying a grasp until it is confident the grasp is stable. We argue that explicitly learning a termination action would be beneficial for many manipulation tasks, since it makes the design of the MDP easier and it forces the policy to truly understand the goal of the task.

\paragraph{Data efficiency} 
Interestingly, our algorithm is more data efficient than the supervised learning based algorithm from ~\citet{levine16} work, achieving higher grasp success with fewer robots continuously generating training data (see Table~\ref{table:sim-data-eff}).

\begin{table}[h]
\begin{center}
\begin{tabular}{ |p{7em}|p{5em}|p{3em}| } 
\hline
Name & Sim Robots &
Success \\
\hline
\multirow{2}{7em}{QT-Opt (ours)} & 30 & 88\% \\
&60 & 95\% \\
\hline
\multirow{3}{7em}{\citet{levine16}} & 60 & 55\% \\
& 280 & 71\% \\
& 1000 & 85\% \\
 \hline
\end{tabular}
\end{center}
\caption{Data efficiency comparison in simulation.}
\label{table:sim-data-eff}
\vspace{-0.25in}
\end{table}
We argue that the algorithm from ~\citet{levine16} is less data efficient because it optimizes a proxy objective of 1-step classification accuracy, which values all data points equally. Our QT-Opt policy values data points based on how they influence reward. This focuses optimization on pivotal decision points that are very important to get right, such as learning when to close the gripper. Doing so lets the model optimize grasp success more efficiently.

\subsection{Effect of Off-Policy Training on Performance}
\label{sec:appendix_off_policy}
\begin{figure*}
\begin{center}
 \includegraphics[width=\textwidth]{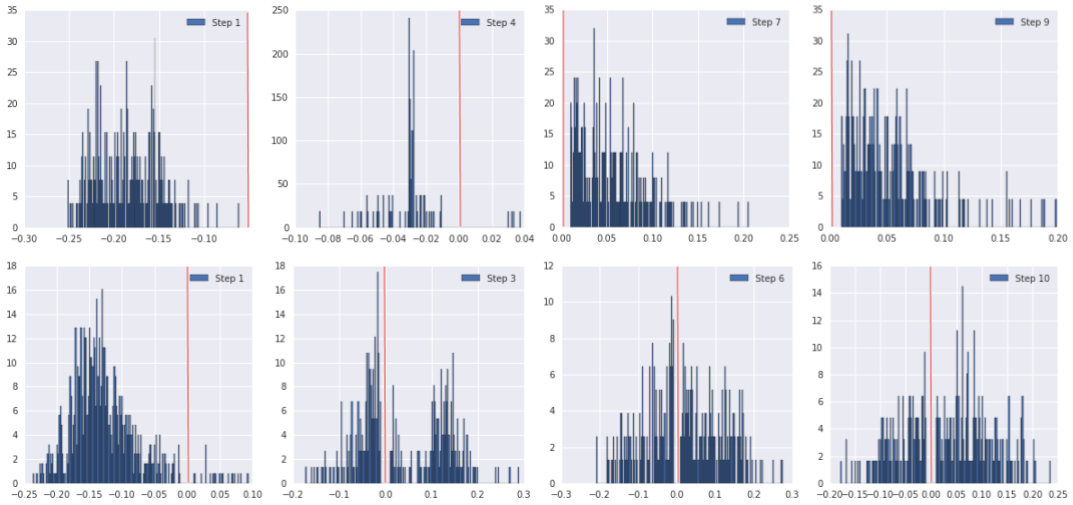}
\end{center}
   \caption{(top row) The distribution of $z$-coordinate of the pose translation of actions selected by the randomized scripted policy. The $z$-coordinate values are biased towards negative values during the first few steps on the descent phase of the episode, then biased towards positive values at later steps on the ascend phase. Such a biased action distribution over states provides poor action space exploration and is insufficient to learn a good Q-Function. (bottom row) The $z$-coordinate of actions selected by a suboptimal QT-Opt policy. The distribution is roughly centered around zero (red axis) at each step providing good action space exploration. }
\label{fig:z_distr}
\vspace{-0.2in}
\end{figure*}
\begin{wraptable}{r}{0.55\textwidth}
\vspace{-0.2in}
\label{table:sim-scripted-explore}
\begin{center}
\begin{tabular}{ |p{7em}|p{7em}|p{3em}| } 
\hline
Num. transitions from \(D_{scripted}\) & Num. transitions from \(D_{explore}\) & Success \\
\hline
300k & 0k & 20\% \\
0k & 300k & 37\% \\
150k & 150k & 86\% \\
200k & 260k & 94\% \\
\hline
\end{tabular}
\end{center}
\caption{Off-policy performance in simulation on datasets with different distributional properties of actions over states}
\label{tbl:sim-distributional-studies}
\end{wraptable}
In principle Q-Learning can learn an optimal policy from off-policy data, as long as the data is sufficiently diverse. Empirically we proved that indeed a decent policy can be trained from our offline dataset. We are interested in answering the following questions: what statistical properties must the data distribution have, and which exploration policies might give rise to that distribution? To that end, we collected two dataset in simulation. The \(D_{scripted}\) dataset was collected by running a randomized scripted policy \(\pi_{scripted}\), discussed in Appendix~\ref{sec:appendix_exploration}, which averaged 30\% grasp success. The second dataset \(D_{explore}\) was collected by running a suboptimal QT-Opt policy \(\pi_{eval}\) which is also averaged 30\% grasp success. Table~\ref{tbl:sim-distributional-studies} shows off-policy performance on different amounts of data sampled from \(D_{scripted}\) and \(D_{explore}\).
This experiment shows that given 300k transitions, it is impossible to learn solely from the \(D_{scripted}\) data, or solely from the \(D_{explore}\) data,
but it is possible to learn with data from both distributions. This shows the importance of collecting off-policy data from a mix of policies with different behavior. The \(\pi_{scripted}\) is a special randomized scripted policy which samples actions in a very biased way towards successful grasps for initial bootstrapping. As a result, it explores state-action space very poorly, yielding only specific greedy actions for certain states. In contrast, a suboptimal \(\pi_{eval}\) makes a lot of suboptimal actions thus resulting in good exploration of the action space. Fig.~\ref{fig:z_distr} visualizes the distribution of the $z$-coordinate of the action translation in the two datasets, showing that $\pi_{eval}$'s actions have less downward bias. We saw similar differences in distributions over other components of the action, like the gripper action. In our real world experiments, early policies had very poor success, as the data was coming primarily from \(\pi_{scripted}\), hence resulting in poor exploration of the action space. However, once we collected enough sufficiently random data, we started getting much better grasp success. We expect this pattern to hold for other real world tasks using QT-Opt: initial performance might be low, but continued data collection should eventually give rise to an appropriately diverse data distribution resulting in a good policy.

\section{Grasping MDP: State Space, Action Space, and Reward Evaluation}
\label{sec:appendix_state_action_reward}
The goal in designing the MDP is to provide a framework in which an agent may learn the necessary hand-eye coordination to reach, grasp, and lift objects successfully. Our sensors include a 640x512 RGB camera and joint position sensors in arm and gripper. In the following we describe representation of state, action and reward and discuss the stopping criterion of our grasping task.

\subsection{Observation Space}
To provide image observations $I_t$ that capture maximal information about the state of the environment, we mounted the camera to the shoulder of the robot overlooking the entire workspace (see Fig.~\ref{fig:robot_setup_and_objects}).
In practice, we also observed that our agent was able to learn more effectively when observations also include some proprioceptive state. Specifically, this state includes a binary open/closed indicator of gripper aperture and the scalar height of the gripper above the bottom of the tray. The full observation is be defined as $\bs_t = (I_t, g_{\text{aperture},t}, g_{\text{height},t})$. The model-input $I_t$ is a 472x472 crop of the full-size image with random crop anchor. This way the model is discouraged from relying on a rigid camera-to-base transformation which cannot be assumed across several robots. Instead, the policy is forced to learn about the presence and the view of the gripper and impact of the actions in the environment, which is critical for the emergence of closed-loop self-corrective behaviours. We also apply image augmentation. The brightness, contrast, and saturation are adjusted by sampling uniformly from $[-0.125, 0.125], [0.5, 1.5]$, and $[0.5, 1.5]$ respectively. To stay consistent, the same augmentation is applied to the images in the current state $\bs$ and next state $\bs'$. Random cropping and image augmentation is only done at train time. At inference time, no augmentation is used and we always crop to the center 472x472 square of the image.

\subsection{Action Space}

The agent's action comprises a gripper pose displacement, and an open/close command. The gripper pose displacement is a difference between the current pose and the desired pose in Cartesian space, encoded as translation $\mathbf{t}_t \in \reals^3$, and vertical rotation encoded via a sine-cosine encoding $\mathbf{r}_t \in \reals^2$. A gripper open/close command is encoded as one-hot vector $[g_{\text{close},t}, g_{\text{open},t}] \in \{0, 1\}^2$. In addition, as this is an episodic task with a finite horizon, our agent has to decide when to terminate. For a baseline policy, we implement a heuristic stopping criterion that triggers when the arm is holding an object above a threshold height. Such a heuristic is task-specific and might bias learning in a sub-optimal way. Thus, we introduce an additional action component $e_t$ which allows our policy to learn a stopping criterion and decide when to terminate autonomously. The full action is defined as $\ba_t = (\mathbf{t}_t, \mathbf{r}_t, g_{\text{close},t}, g_{\text{open},t}, e_t)$.\\

\subsection{Reward Function}
\begin{wrapfigure}{r}{0.5\textwidth}
\vspace{-0.5cm}
\begin{center}
\includegraphics[width=0.5\textwidth]{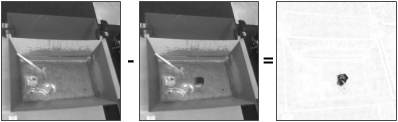}
\end{center}
\vspace{-0.35cm}
\caption{Grasp success is determined by subtracting images before an object is dropped into the bin (left) and after it was dropped (right).}
\label{fig:my_label}
\vspace{-0.2cm}
\end{wrapfigure}

The agent is given a reward of 1 at the end of an episode if it has successfully grasped an object from the bin. In all other cases the reward is 0. In order to detect a successful grasp we use an image subtraction test, see Fig.~\ref{fig:my_label}. An image of the scene is captured after the grasp attempt with the arm moved out of camera view. Then a second image is taken after attempting to drop the grasped object into the bin. If no object was grasped, these two images would be identical. Otherwise, if an object was picked up, certain pixels in the two images will be different. A grasp is labeled success if the image subtraction is above a threshold.

Because we use a simple image subtraction test to compute the reward, in practice the labeling is not perfect. We analyzed the failed grasps from the model that reaches 96\% grasp success, and discovered that out of 28 misgrasps, 3 were actually successful grasps that have been classified improperly by our image subtraction test. Our background subtraction test might not detect small objects, making it vulnerable to false negative registrations. This shows that our learning algorithm can train a very good policy even with a small noise in the reward function.

\subsection{Grasp execution and termination condition}
\label{appendix:termination}
\begin{algorithm}[h]
{\small
	\caption{Grasping control-loop}
	
	\label{alg:servo}
	\begin{algorithmic}[1]
	\STATE Pick a policy \(policy\).
	\STATE Initialize a \(robot\).
	\WHILE{$step<N$ and not {\it terminate\_episode}}
    \STATE $\bs$ = \(robot\).CaptureState()
    \STATE $\ba$ = \(policy\).SelectAction($\bs$)
    \STATE \(robot\).ExecuteCommand($\ba$)
    \STATE {\it terminate\_episode} = $e$ \COMMENT{Termination action $e$ is either learned or decided heuristically.}
    \STATE r = \(robot\).ReceiveReward()
    \STATE emit($\bs, \ba,$ r)
   \ENDWHILE
 	\end{algorithmic}
}
\end{algorithm}
With state, action and reward defined in the sections above, we now describe the control-loop executed on our robots. As described in Algorithm~\ref{alg:servo} we control our manipulator by closing a loop around visual input and gripper commands. At the end of an episode we detect grasp success and return the sequence of state-action pairs to be fed back into the learning pipeline or stored offline. Depending on the executed \(policy\), our termination action $e$ is either learned or decided heuristically as we discuss below.
\paragraph{Scripted termination condition} Initial experiments used a heuristic termination condition to determine when the policy is done. The heuristic detects when the gripper is holding an object, and the policy is continuing to command upward actions after it has passed a height threshold.

\begin{algorithm}[h]
{\small
	\caption{Scripted termination condition}
	\label{alg:scripted_termination_condition}
	\begin{algorithmic}[1]
	    \STATE termination\_height = 0.13
		\STATE s = robot.GetState()
		\STATE gripper\_height = s.height\_of\_gripper
		\STATE gripper\_closed = s.gripper\_status='CLOSED'
		\STATE a = robot.GetAction(s)
		\STATE next\_action\_height = a.translation.z
		\IF{$gripper\_closed\ and\ gripper\_height>termination\_height\ and\ next\_action\_height>gripper\_height$}
			\STATE terminate = True
		\ELSE
			\STATE terminate = False
		\ENDIF
	\end{algorithmic}
}
\end{algorithm}

\paragraph{Learned termination condition}
\label{sec:grasp_success}
In principle, a policy may use the camera image to determine if it has successfully grasped and raised an object. It may inform such knowledge through a discrete termination action. To learn the termination action, the reward structure must be slightly changed, \(reward=1\) is assigned when \(grasp\_success\ = True\ and\ gripper\_termination\_height>threshold\). This enforces the policy to indicate episode termination only after the gripper lifts the object, which makes it easier to provide visual feedback for robust grasps; objects will fall out of a brittle grasp during the lifting process, and terminating the grasp after the object falls gives richer negative examples.

\section{Q-Function Neural Network Architecture}
\label{sec:appendix_arch}
\begin{figure*}
\begin{center}
 \includegraphics[width=\textwidth]{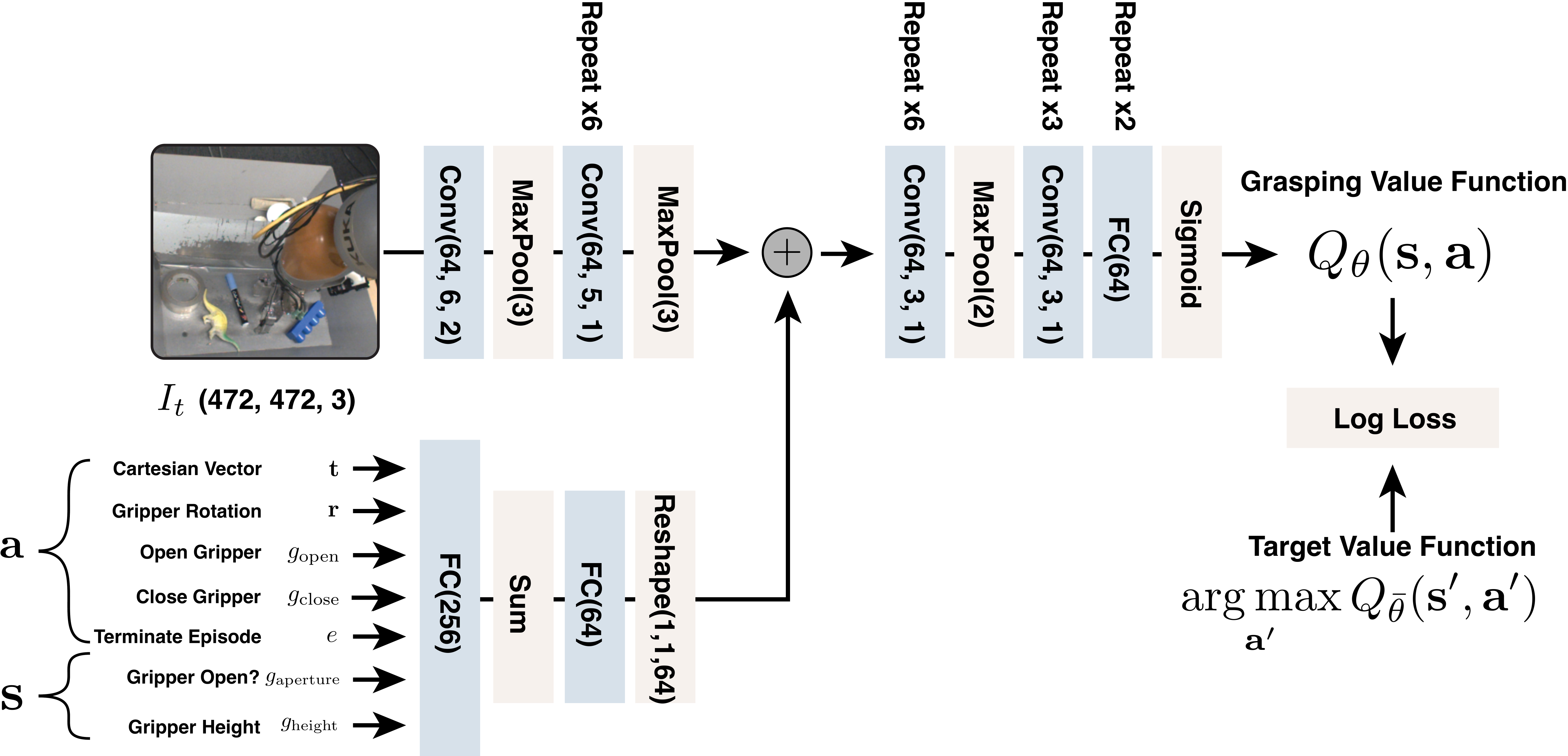}
\end{center}
   \caption{The architecture of grasping Q-Function. The input image is processed by a stack of convolutional layers before the introduction of the action and vector-valued state variables (gripper status and height). These are processed by several fully connected layers, tiled over the width and height dimension of the convolutional map, and added to it. The resulting convolutional map is further processed by a number of convolutional layers until the output. The output is gated through a sigmoid, such that our Q-values are always in the range $[0, 1]$.}
\label{fig:q-network}
\vspace{-0.2in}
\end{figure*}

We model the Q-function as a large deep neural network whose architecture, shown in Figure \ref{fig:q-network}, is inspired by the one presented in~\cite{levine16}. The network takes the monocular RGB image component of the state $\bs$ as input, and processes it with 7 convolutional layers. We transform actions $\ba$ and additional state features ($g_\text{aperture}$, $g_\text{height}$) with fully-connected layers, then merge them with visual features by broadcasted element-wise addition. After fusing state and action representations, the Q value $Q_\theta(\bs, \ba)$ is modeled by 9 more convolution layers followed by two fully-connected layers. Each convolution and fully-connected layer uses batch normalization~\cite{is-bnad-15} and the ReLU nonlinearity. The value function handles mixtures of discrete actions (open, close, terminate) and continuous actions (Cartesian vector and gripper rotation), making the same state-action space amenable to learning future tasks like placing and stacking. In Table~\ref{table:state-repr-ablation}, we demonstrate empirically that the model performs better when provided with redundant state representations ($g_\text{aperture}$, $g_\text{height}$). All weights are initialized with truncated normal random variables ($\sigma=.01$) and L2-regularized with a coefficient of $7\mathrm{e}{-5}$. Models are trained with SGD with momentum, using learning rate $0.0001$ and momentum $0.9$.

\section{QT-Opt Distributed Reinforcement Learning System Design}
\label{sec:appendix_distributed_rl_infra}
\begin{figure*}
\begin{center}
 \includegraphics[width=\textwidth]{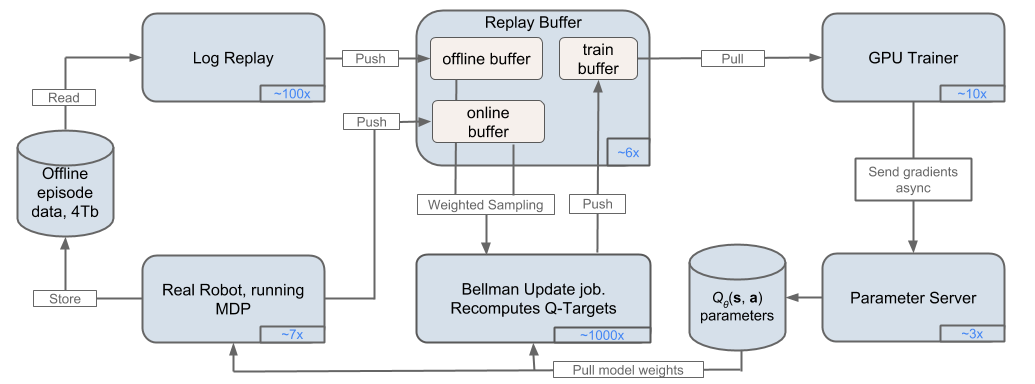}
\end{center}
   \caption{Architecture of the QT-Opt distributed reinforcement learning algorithm.}
\label{fig:distributed_infra}
\vspace{-0.2in}
\end{figure*}

Prior works~\citep{stooke_accelerated,impala} have explored new reinforcement learning algorithms that allow compute parallelization to accelerate training. We propose a parallel and distributed learning algorithm, inspired by~\citep{gorila}, that can scale easily on cloud-based infrastructures. Fig.~\ref{fig:distributed_infra} is a diagram of the full distributed system.

This distributed design of our QT-Opt algorithm was important for several reasons: 
\begin{enumerate}[(a)]
\item %
We use high-resolution images. Trying to store all transitions in the memory of a single machine is infeasible. Thus we employ a distributed replay buffer, which lets us store hundreds of thousands of transitions across several machines.
\item %
The Q-network is quite large, and distributing training across multiple GPUs drastically increases research velocity by reducing time to convergence. Similarly, in order to support large scale simulated experiments, the design has to support running hundreds of simulated robots that cannot fit on a single machine.
\item %
Decoupling training jobs from data generation jobs allows us to treat training as data-agnostic, making it easy to switch between simulated data, off-policy real data, and on-policy real data. It also lets us scale the speed of training and data generation independently.
\end{enumerate}

Similar to~\citep{horgan2018distributed}, our distributed replay buffer is shared across all agents, although we do not use prioritized replay. 
We distribute training across 10 GPUs, using asynchronous SGD with momentum as proposed by~\citep{dean_async}. Models were trained using  TensorFlow~\citep{tensorflow2015-whitepaper}. This system allows us to train the Q-function at 40 steps per second with a batch size of 32 across 10 NVIDIA P100 GPUs.

\subsection{Online Training}
\label{sec:online-training}

Online agents (real or simulated robots) collect data from the environment. The policy used is the Polyak averaged weights $Q_{\bar{\theta}_1}(s,a)$ and the weights are updated every 10 minutes. That data is pushed to a distributed replay buffer, which we call the \enquote{online buffer}. The data is also persisted to disk for future offline training.

\subsection{Offline Training from Logs}

To support offline training, we run a log replay job. This job reads data sequentially from disk for efficiency reasons. It replays saved episodes as if an online agent had collected that data. This lets us seamlessly merge off-policy data with on-policy data collected by online agents as described in Appendix~\ref{sec:policy_fine_tuning}.

Offline data comes from all previously run experiments. In total, we collected \numgrasps{} grasps of offline data, which took 4 terabytes of disk space.
 In fully off-policy training, we can train the policy by loading all data with the log replay job, letting us train without having to interact with the real environment.

Despite the scale of the distributed replay buffer, we still can't fit the entire dataset into memory. In order to be able to visit each datapoint uniformly, we keep continuously running the Log Replay to refresh the in-memory data residing in the Replay Buffer. In practice we found that running at least 100 Log Replay jobs was important to mitigate correlations between consecutive episodes and drive sufficiently diverse data to the Replay Buffer.

\subsection{Online Joint Finetuning After Offline Training}
\label{sec:policy_fine_tuning}

In practice, we begin with off-policy training to initialize a good policy, and then switch to on-policy joint finetuning. To do so, we first train fully off-policy by using the Log Replay job to replay episodes from prior experiments. After training off-policy for enough time, we restart QT-Opt, training with a mix of on-policy and off-policy data. We train off-policy for 5M to 15M steps.

Real on-policy data is generated by 7 KUKA LBR IIWA robot arms where the weights of the policy $Q_{\bar{\theta}_1}(s,a)$ are updated every 10 minutes. Compared to the offline dataset, the rate of on-policy data production is much lower and the data has less visual diversity. However, the on-policy data also contains real-world interactions that illustrate the faults in the current policy. To avoid overfitting to the initially scarce on-policy data, we gradually ramp up the fraction of on-policy data from 1\% to 50\% over the first 1M gradient update steps of joint finetuning training.

Since the real robots can stop unexpectedly (e.g., due to hardware faults), data collection can be sporadic, potentially with delays of hours or more if a fault occurs without any operator present. This can unexpectedly cause a significant reduction in the rate of data collection. To mitigate this, on-policy training is also gated by a \textit{training balancer}, which enforces a fixed ratio between the number of joint finetuning gradient update steps and number of on-policy transitions collected. The ratio was defined relative to the speed of the GPUs and of the robots, which changed over time. We did not tune this ratio very carefully, and in practice the ratio ranged from 6:1 to 11:1.

\subsection{Distributed Bellman Update}
\label{sec:appendix_bellman_update}
To stabilize deep Q-Learning, we use a target network, as discussed in Section~\ref{sec:qtopt}. Since target networks parameters typically lag behind the online network when computing TD error, the Bellman backup can actually be performed asynchronously in a separate process. We propose a novel algorithm that
computes ${r(\bs,\ba) + \gamma V(\bs')}$
in parallel on separate CPU machines, storing the output of those computations in an additional buffer named \enquote{train} in our distributed replay buffer. We call this job the Bellman updater, and have been able to scale this computation up to 1,000 machines using around 14k cores.

Let $\theta$ be the parameters of the current Q-network, $\bar{\theta}$ be the parameters of the target Q-network, and $p(\bs,\ba,\bs)$ be the distribution of transitions in the replay buffer. As discussed in ~\ref{sec:rlql}, our Bellman backup is formulated as:

\[
\bellman(\theta) = \E_{(\bs,\ba,\bs') \sim p(\bs,\ba,\bs')} \left[ D \left(
Q_\theta(\bs,\ba) , r(\bs,\ba) + \gamma V(\bs') \right)
\right].
\]

Note that because we use several Bellman updater replicas, each replica will load a new target network at different times. All replicas push the Bellman backup to the shared replay buffer in the \enquote{train buffer}. This makes our target Q-values effectively generated by an ensemble of recent target networks, sampled from an implicit distribution $R_t$. Expanding the clipped Double DQN estimation of value gives the following objective:

\begin{equation}
\bellman(\theta) = \E_{(\bs,\ba,\bs') \sim p(\bs,\ba,\bs')} \left[ \E_{(\bar{\theta}_1,\bar{\theta}_2) \sim R_t(\bar{\theta}_1,\bar{\theta}_2)} \left[ D \left(
Q_\theta(\bs,\ba) , r(\bs,\ba) + \gamma V_{\bar{\theta}_1, \bar{\theta}_2}(\bs') \right)
\right] \right]
\end{equation}

where $V_{\bar{\theta}_1,\bar{\theta}_2}(\bs')$ is estimated using clipped Double DQN:

\begin{equation*}
    V_{\bar{\theta}_1,\bar{\theta}_2}(\bs') = \min_{i=1,2} Q_{\bar{\theta}_i}(\bs',\arg\max_{\ba'} Q_{\bar{\theta}_1}(\bs',\ba'))
\end{equation*}

Even if a Q-function estimator is unbiased, variance in TD error estimates are converted into over-estimation bias during the Bellman backup (via the max operator). We hypothesize that ensembling via a distribution of lagging target networks stabilizes training by reducing the variance (and thereby reduces bias in target values). We also decrease overestimation bias by using Clipped Double-Q Learning and a damped discount factor of $\gamma=.9$. We found this training to be stable and reproducible. The target network $Q_{\bar{\theta}_1}$ is computed by doing a Polyak averaging ~\citep{polyak1992acceleration} with a decay factor of $0.9999$, while $Q_{\bar{\theta}_2}$ is lagged on average by about 6000 gradient steps. The exact amount of delay varies because of the asynchronous nature of the system.

\subsection{Distributed Replay Buffer}

The distributed replay buffer supports having named replay buffers. We create three named buffers: \enquote{online buffer} holds online data, \enquote{offline buffer} holds offline data, and \enquote{train buffer} stores Q-targets computed by the Bellman updater. The replay buffer interface supports weighted sampling from the named buffers, which is useful when doing on-policy joint finetuning. The distributed replay buffer is spread over 6 workers, which each contain $50k$ transitions, totalling $6*50k=300k$ transitions. All buffers are FIFO buffers where old values are removed to make space for new ones if the buffer is full.

\end{document}